\documentclass[a4paper,fleqn]{cas-sc}

\usepackage[authoryear]{natbib}
\usepackage{pifont}
\usepackage{enumitem}
\usepackage{array}  

\usepackage[table]{xcolor}
\usepackage{tabularx}
\usepackage{booktabs}
\usepackage{caption}

\def\tsc#1{\csdef{#1}{\textsc{\lowercase{#1}}\xspace}}
\tsc{WGM}
\tsc{QE}
\tsc{EP}
\tsc{PMS}
\tsc{BEC}
\tsc{DE}

\begin{document}
\let\WriteBookmarks\relax
\def\floatpagepagefraction{1}
\def\textpagefraction{.001}

\shorttitle{Cooperative World-Action Modeling for End-to-End Autonomous Driving}

\shortauthors{Junwei You et~al.}

\title [mode = title]{V2X-WAM: A Cooperative World Action Model for End-to-End Autonomous Driving}                      

\author[1,2]{Junwei You}[orcid=0009-0002-6447-8276]
\author[3]{Weizhe Tang}
\author[4,5]{Can Wang}\cormark[1]
\author[6]{Yan Zhao}
\author[1,2]{Jun Hua}
\author[7]{Haotian Shi}
\author[1,2]{Wei Zhang}
\author[1,2]{Lin Wang}\cormark[1]
\author[3]{Bin Ran}





\affiliation[1]{organization={ITS Center, Research Institute of Highway Ministry of Transport},
    city={Beijing},
    postcode={100088}, 
    country={China}}

\affiliation[2]{organization={State Key Lab of Intelligent Transportation System, Research Institute of Highway Ministry of Transport},
    city={Beijing},
    postcode={100088}, 
    country={China}}


\affiliation[3]{organization={Department of Civil and Environmental Engineering, University of Wisconsin–Madison},
    city={Madison},
    state={WI},
    postcode={53706}, 
    country={USA}}

\affiliation[4]{
    organization={Intelligent Transportation Systems Research Center, Wuhan University of Technology},
    city={Wuhan},
    postcode={430063},
    country={China}
}

\affiliation[5]{
    organization={Engineering Research Center of Transportation Information and Safety, Ministry of Education},
    city={Wuhan},
    postcode={430063},
    country={China}
}

\affiliation[6]{
    organization={School of Transportation, Inner Mongolia University},
    city={Hohhot},
    postcode={010021},
    country={China}
}
    
    
\affiliation[7]{organization={College of Transportation, Tongji University},
    city={Shanghai},
    postcode={201804}, 
    country={China}}








\cortext[cor1]{Corresponding author}




\begin{abstract}
Vehicle--infrastructure cooperation can complement onboard sensing with broader and more informative observations of the traffic environment, providing valuable support for end-to-end autonomous driving. However, existing cooperative driving methods mainly exploit roadside information to enhance the representation of the current scene, while the future consequences of prospective driving actions are rarely modeled explicitly. This limits the ability of the planner to anticipate how its decisions may interact with the evolving traffic environment. To address this issue, we propose V2X-WAM, a cooperative world action model that tightly couples cooperative scene understanding, action generation, and future-world reasoning. V2X-WAM constructs a reliability-aware spatiotemporal representation from vehicle- and infrastructure-side observations, while compressing infrastructure information into a compact quantized message for efficient communication. Based on the resulting cooperative representation, a multimodal planner generates prospective trajectories, which explicitly condition future occupancy and dynamic-flow prediction. The predicted world consequences are then fed back to refine the planned trajectory, forming a closed interaction between action and future-world evolution. Experiments on a large-scale real-world cooperative driving dataset demonstrate that V2X-WAM consistently improves planning accuracy and safety over representative end-to-end cooperative driving methods, while achieving stronger future-world prediction and substantially lower communication overhead. Ablation studies further validate the effectiveness of the proposed design.
\end{abstract}

\begin{keywords}
End-to-end autonomous driving \sep Vehicle--infrastructure cooperation \sep World model \sep World--action modeling \sep Communication-efficient V2X
\end{keywords}

\maketitle

\section{Introduction}
\label{sec:introduction}

End-to-end autonomous driving has rapidly evolved from direct sensor-to-trajectory prediction toward multimodal driving models with increasingly stronger semantic understanding and reasoning capabilities. Large language models (LLMs) and vision-language models (VLMs) have enabled driving systems to jointly interpret visual observations, navigation instructions, and complex driving context. LMDrive \citep{shao2024lmdrive}, for example, incorporates language instructions into closed-loop autonomous driving, while EMMA \citep{hwang2025emma} develops a generalist multimodal model that jointly supports trajectory planning and other driving-related tasks. Building on these advances, vision-language-action (VLA) models further integrate multimodal reasoning with action generation. AutoVLA \citep{zhou2025autovla} combines reasoning and trajectory generation through explicit action modeling, while OpenDriveVLA \citep{zhou2026opendrivevla} jointly aligns visual representations, language instructions, and spatial driving actions. By strengthening the connection between scene understanding and action generation, these models provide a more unified perception--reasoning--action framework. Nevertheless, their decisions are still predominantly conditioned on the currently observed scene, without explicitly reasoning about how the traffic environment may evolve after a prospective action is executed.

\begin{figure}[pos=htbp]
    \centering
    \includegraphics[width=\textwidth]{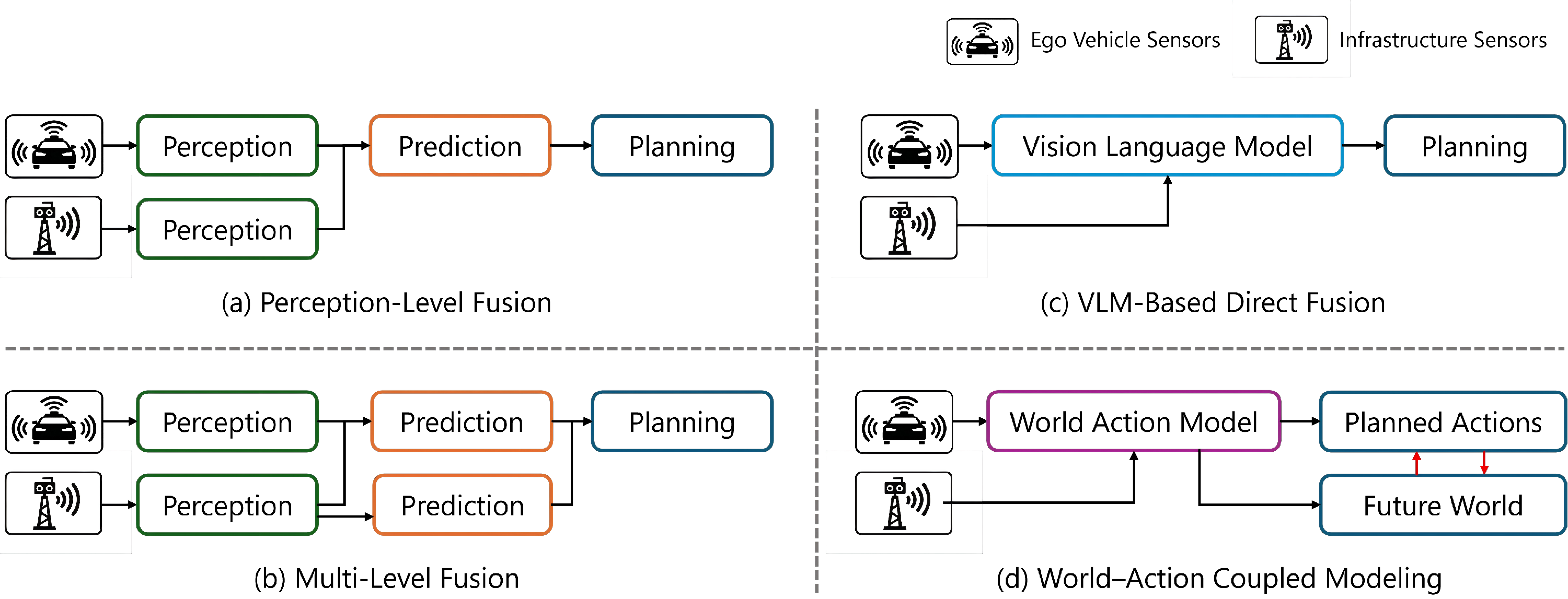}
    \caption{Evolution of end-to-end cooperative autonomous driving paradigms.}
    \label{fig:paradigm}
\end{figure}

World model-based autonomous driving introduces this prospective dimension by modeling future environmental evolution as part of the decision process. Drive-WM \citep{wang2024drivewm} generates controllable future observations under different driving maneuvers, while LAW \citep{li2025law} predicts future latent scene representations conditioned on ego trajectories. World4Drive \citep{zheng2025world4drive} further employs intention-conditioned latent futures to support multimodal trajectory generation and evaluation. More recent studies have tightened the interaction between future modeling and action generation. LDrive \citep{tan2026ldrive} interleaves action proposals with latent representations of their possible outcomes, while Latent-WAM \citep{wang2026latentwam} jointly models latent world dynamics and driving actions within a unified framework. These developments move autonomous driving from reactive action generation toward consequence-aware decision making, where prospective actions are explicitly related to future-world evolution. However, existing world action models remain predominantly ego-centric, and both action generation and future-world reasoning are therefore constrained by what onboard sensors can observe.

Vehicle-to-everything (V2X) cooperative driving provides a direct means of extending this observation boundary by allowing the ego vehicle to exploit complementary information from connected vehicles and roadside infrastructure. Cooperative sensing can reveal occluded traffic participants, extend the observable range, and provide additional spatial and temporal context that is unavailable from onboard sensing alone. Within this broader paradigm, recent research has increasingly progressed toward end-to-end cooperative autonomous driving, in which shared information contributes directly to prediction and planning. As illustrated in Fig.~\ref{fig:paradigm}, end-to-end cooperative autonomous driving has evolved from perception-level fusion toward multi-level integration and, more recently, direct multimodal reasoning. UniV2X \citep{yu2025univ2x} integrates cooperative detection, mapping, occupancy prediction, and planning within a unified end-to-end framework, enabling infrastructure information to influence the final driving trajectory through multiple intermediate tasks. UniMM-V2X \citep{song2026unimmv2x} further extends cooperation across different stages of the driving process to strengthen the contribution of shared information to downstream planning. Moving beyond predefined intermediate task interfaces, V2X-VLM \citep{you2026v2xvlm} introduces vision-language reasoning into cooperative end-to-end driving and directly connects vehicle- and infrastructure-side observations with trajectory planning. More recent methods such as OmniV2X \citep{peng2026omniv2x} and DH-VLM \citep{song2026dhvlm} continue this progression through generative planning and cooperative latent reasoning. These developments have substantially tightened the connection between cooperative observations and driving actions. However, existing end-to-end cooperative methods still lack an explicit coupling between action generation and future-world evolution: cooperative information informs the planned trajectory, but the planned action does not explicitly condition the predicted future world, nor are the predicted consequences subsequently used to refine that action.

To address this limitation, we propose V2X-WAM, a cooperative world action model for end-to-end autonomous driving, which extends world-action reasoning to the cooperative driving setting. V2X-WAM jointly exploits vehicle- and infrastructure-side observations to support both action generation and future-world modeling. Temporal multimodal observations from the ego vehicle and infrastructure are first encoded into a shared cooperative scene representation. To account for the heterogeneous quality of cooperative information, infrastructure features are transmitted through a compact quantized message and adaptively regulated by learned spatial reliability and temporal confidence. The resulting cooperative representation is compressed into shared scene latents, from which the planning branch generates multimodal trajectory proposals and obtains a preliminary action sequence. These planned actions then condition future-world prediction, where future cooperative occupancy and dynamic flow characterize the spatial evolution and motion of surrounding traffic. The predicted world consequences are subsequently fed back to the planning branch to refine the preliminary trajectory, forming an explicit action-to-world-to-action interaction. Through this coupling, V2X-WAM enables cooperative observations to support not only a more complete representation of the current traffic scene, but also consequence-aware reasoning over how the scene may evolve in response to the ego vehicle's prospective actions.

The main contributions of this work are summarized as follows:
\begin{itemize}
    \item We propose V2X-WAM, a cooperative world-action framework for end-to-end autonomous driving that explicitly couples action generation with future-world modeling. Planned actions condition future traffic evolution, while predicted world consequences are fed back to refine the final trajectory.

    \item We develop a reliability-aware cooperative scene representation that integrates temporal vehicle- and infrastructure-side observations through a compact quantized infrastructure message. Learned spatial reliability and temporal confidence jointly regulate the contribution of cooperative information before it is encoded into the shared latent representation for planning and future-world modeling.

    \item We introduce an action-conditioned future-world modeling mechanism that jointly predicts future occupancy and dynamic flow, together with flow-guided occupancy transport to capture the spatial and dynamic evolution of surrounding traffic.
    
    \item We conduct comprehensive experiments on a large-scale real-world cooperative driving dataset, demonstrating the effectiveness of V2X-WAM, with additional ablation studies further examining the contributions of its key components.
\end{itemize}

The remainder of this paper is organized as follows. Section~\ref{sec:related_work} reviews related studies on end-to-end autonomous driving, world models, and cooperative autonomous driving. Section~\ref{sec:methodology} presents the proposed V2X-WAM framework and its main components. Section~\ref{sec:experiments} describes the experimental setup and reports the quantitative, ablation, and qualitative results. Finally, Section~\ref{sec:conclusion} concludes the paper and discusses future research directions.

\section{Related Work}
\label{sec:related_work}

\subsection{End-to-End Autonomous Driving with Foundation Models and World Models}

Recent end-to-end autonomous driving research has increasingly incorporated foundation models to enhance scene understanding, reasoning, and planning. Early language-based approaches explored the use of large language models for trajectory generation and closed-loop decision making. LMDrive \citep{shao2024lmdrive} integrates language instructions with multimodal observations for closed-loop autonomous driving, while DriveVLM \citep{tian2025drivevlm} employs a large vision-language model for scene description, reasoning, and hierarchical planning. EMMA \citep{hwang2025emma} further formulates multiple driving tasks, including trajectory planning, perception, and road-graph prediction, within a unified multimodal language-model framework. Other studies combine high-level language reasoning with specialized driving policies. Senna \citep{jiang2024senna}, for example, separates vision-language-based high-level planning from precise trajectory prediction, while OmniDrive \citep{wang2025omnidrive} strengthens 3D-aware reasoning and planning through counterfactual supervision. Building upon these advances, vision-language-action models more explicitly incorporate the action space into multimodal modeling. AutoVLA \citep{zhou2025autovla} tokenizes feasible trajectories and unifies semantic reasoning with action generation in an autoregressive model, whereas OpenDriveVLA \citep{zhou2026opendrivevla} models structured agent--environment--ego interactions during trajectory decoding. Recent methods further improve action specialization and language--action alignment: DriveMoE \citep{yang2026drivemoe} introduces scene- and skill-specialized experts for perception and action generation, while LinkVLA \citep{wang2026linkvla} establishes a shared representation between language and actions to improve alignment and inference efficiency. These approaches progressively integrate perception, reasoning, and action generation, but their driving policies are still primarily learned from the relationship between current observations and future actions.

World models complement this action-oriented paradigm by explicitly learning how driving scenes evolve over time. Drive-WM \citep{wang2024drivewm} employs a generative world model to produce controllable multiview future observations under different driving maneuvers and evaluates candidate plans through the generated futures. Rather than generating future images, LAW \citep{li2025law} learns future latent scene representations conditioned on ego trajectories, using future prediction as self-supervision for end-to-end planning. World4Drive \citep{zheng2025world4drive} further constructs intention-aware latent future states and uses the learned world representation to generate and evaluate multimodal trajectories. More recently, DriveLaW \citep{xia2026drivelaw} unifies video generation and motion planning by directly transferring predictive latent representations from a driving world model to a diffusion planner. These studies demonstrate that future-world representations can provide informative supervision and evaluation signals for trajectory planning, shifting end-to-end driving from purely reactive policies toward predictive decision making.

A growing line of work further integrates future modeling with action generation rather than treating the two as separate objectives. LDrive \citep{tan2026ldrive} performs latent chain-of-thought reasoning by interleaving action-proposal tokens with world-model tokens representing the possible outcomes of those actions. VLA-World \citep{wang2026vlaworld} uses an action-derived trajectory to guide future visual imagination and subsequently reasons over the generated future to refine planning. Latent-WAM \citep{wang2026latentwam} develops spatially aware compressed scene representations together with a dynamic latent world model for end-to-end trajectory planning. Recent world-action models have also explored more efficient coupling strategies: Metis \citep{li2026metis} employs separate video and action experts within a unified world-action framework, whereas SimWAM \citep{zhao2026simwam} uses video generation as a training signal and retains only the action branch during inference. Collectively, these works establish increasingly direct connections between action generation and modeled world dynamics. However, the underlying world representations are predominantly constructed from ego-vehicle observations, leaving the resulting world-action reasoning constrained by the spatial visibility and information coverage of onboard sensors.

\subsection{End-to-End Cooperative Autonomous Driving}

Cooperative autonomous driving extends end-to-end policies by allowing driving decisions to exploit complementary information exchanged among connected vehicles and roadside infrastructure. COOPERNAUT \citep{cui2022coopernaut} is an early end-to-end cooperative driving framework that communicates compact LiDAR representations among connected vehicles and directly learns driving policies from aggregated cooperative information. CoDriving \citep{liu2025codriving} subsequently introduces driving-oriented communication to selectively transmit decision-relevant features and integrates collaboration across the driving pipeline. Moving toward a more complete vehicle--infrastructure architecture, UniV2X \citep{yu2025univ2x} jointly incorporates cooperative detection, online mapping, occupancy prediction, and trajectory planning through sparse--dense hybrid communication. These studies establish the feasibility of optimizing cooperative representations directly for downstream driving rather than treating V2X information solely as an auxiliary perception input.

Subsequent research has expanded cooperation across different reasoning levels and modalities. UniMM-V2X \citep{song2026unimmv2x} introduces multi-level cooperation across perception, prediction, and planning, together with mixture-of-experts modules for task-adaptive representation learning. V2X-VLM \citep{you2026v2xvlm} further extends cooperative driving toward multimodal semantic reasoning by combining vehicle- and infrastructure-side observations with vision-language representations for end-to-end trajectory planning. Building on this direction, SEAL \citep{you2026seal} focuses on safe and robust cooperative driving under long-tail conditions through adaptive multimodal modeling, incorporating scenario-aware feature adaptation and contrastive learning to improve reasoning and planning under rare and visually degraded environments. V2X-UniPool \citep{luo2026v2xunipool} further transforms heterogeneous V2X observations into a unified knowledge pool and employs retrieval-augmented reasoning to support planning with reduced communication overhead. These methods broaden cooperative interaction from geometric feature sharing toward semantic, scenario-aware, and knowledge-level representations.

More recent work has incorporated foundation-model reasoning and generative planning into cooperative autonomous driving. OmniV2X \citep{peng2026omniv2x} introduces a generative foundation planner that directly conditions trajectory generation on multimodal multi-agent context and supports efficient adaptation from large-scale single-agent driving data. DH-VLM \citep{song2026dhvlm} adopts asymmetric semantic cooperation, where infrastructure-side global reasoning is compressed into latent guidance and transferred to the ego vehicle for local planning. AURORA \citep{xu2026aurora} further combines cross-view vehicle--roadside alignment, vision-language reasoning, and generative trajectory planning within a closed-loop cooperative framework. These developments progressively move end-to-end cooperation from shared perception toward multimodal reasoning, latent information exchange, and direct trajectory generation. Nevertheless, existing methods still primarily use cooperative information to improve scene representations or condition planning outputs, while explicit action-conditioned cooperative world modeling and subsequent world-to-action feedback remain largely unexplored. V2X-WAM addresses this gap by coupling prospective actions with cooperative future-world evolution and feeding the modeled consequences back into trajectory refinement.

\section{Methodology}
\label{sec:methodology}

\subsection{Problem Formulation and Framework Overview}
\label{sec:overview}

We consider end-to-end cooperative autonomous driving with an ego vehicle and a paired roadside infrastructure unit. At the current time $t$, agent $a\in\{e,i\}$, where $e$ and $i$ denote the ego vehicle and infrastructure, respectively, provides a camera observation $\mathbf{I}_{t}^{a}$ and a LiDAR bird's-eye-view (BEV) history $\mathbf{B}_{t-H+1:t}^{a}$ over $H$ historical steps. All historical BEV observations from both agents are transformed into the coordinate frame of the current ego vehicle, such that temporal evolution and cross-agent information are represented in a common spatial reference.

In addition to multimodal observations, the model receives the historical ego positions $\mathbf{P}_{t}^{\mathrm{hist}}\in\mathbb{R}^{H\times 2}$, the current ego-motion state $\mathbf{s}_{t}$, a high-level route command $r_t$, the relative vehicle--infrastructure geometry $\mathbf{g}_{t}$, and a drivable-area map $\mathbf{M}_{t}^{\mathrm{drv}}$. The complete model input is written as
\begin{equation}
\mathcal{X}_{t}
=
\left\{
\mathbf{I}_{t}^{e},
\mathbf{I}_{t}^{i},
\mathbf{B}_{t-H+1:t}^{e},
\mathbf{B}_{t-H+1:t}^{i},
\mathbf{P}_{t}^{\mathrm{hist}},
\mathbf{s}_{t},
r_t,
\mathbf{g}_{t},
\mathbf{M}_{t}^{\mathrm{drv}}
\right\}.
\label{eq:input}
\end{equation}

Given $\mathcal{X}_{t}$, V2X-WAM predicts the ego trajectory
$\hat{\mathbf{P}}
=
[\hat{\mathbf{p}}_{1},\ldots,\hat{\mathbf{p}}_{T}]
\in\mathbb{R}^{T\times 2}$,
future cooperative occupancy
$\hat{\mathbf{O}}_{1:T}$,
and future dynamic flow
$\hat{\mathbf{V}}_{1:T}$.
Here, $\hat{\mathbf{O}}_{\tau}$ represents surrounding traffic occupancy in the fixed current-ego BEV frame, while
$\hat{\mathbf{V}}_{\tau}$ describes the planar displacement of dynamic actors between consecutive future states.

Rather than predicting planning and future-world states as independent outputs of a shared encoder, V2X-WAM factorizes the driving process as
\begin{equation}
\begin{aligned}
&\mathbf{Z}
=
f_{\mathrm{scene}}(\mathcal{X}_{t}),
\\
&(\hat{\mathbf{P}}^{(0)},\hat{\mathbf{A}},\mathbf{Q})
=
f_{\mathrm{plan}}(\mathbf{Z},\mathbf{s}_{t}),
\\
&(\hat{\mathbf{O}},\hat{\mathbf{V}},\mathbf{W})
=
f_{\mathrm{world}}
\left(
\mathbf{Z},
\hat{\mathbf{P}}^{(0)},
\hat{\mathbf{A}}
\right),
\\
&\hat{\mathbf{P}}
=
f_{\mathrm{ref}}
\left(
\hat{\mathbf{P}}^{(0)},
\mathbf{Q},
\mathbf{W}
\right).
\end{aligned}
\label{eq:wam_factorization}
\end{equation}
In Eq.~\eqref{eq:wam_factorization},
$\mathbf{Z}$ denotes the shared cooperative scene latents,
$\hat{\mathbf{P}}^{(0)}$ is the preliminary trajectory,
$\hat{\mathbf{A}}$ is the corresponding prospective action sequence,
$\mathbf{Q}$ contains the selected planning tokens, and
$\mathbf{W}$ denotes the predicted future-world consequence tokens.
The formulation establishes an explicit action-to-world-to-action interaction: the preliminary action conditions future-world evolution, and the predicted consequences subsequently refine that action.

\begin{figure}[pos=htbp]
    \centering
    \includegraphics[width=0.9\textwidth]{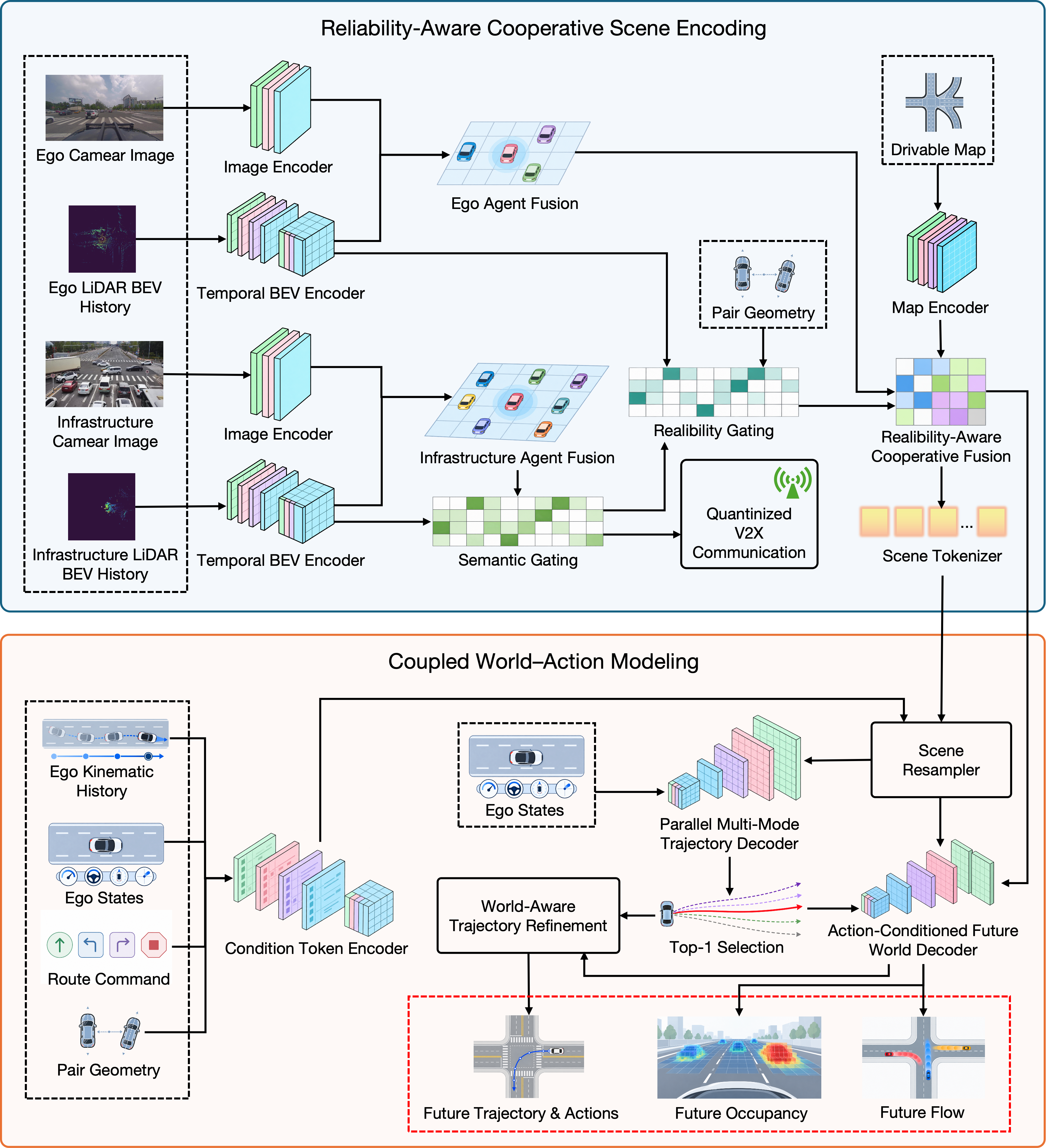}
    \caption{Overall architecture of the proposed V2X-WAM.}
    \label{fig:framework}
\end{figure}

As illustrated in Fig.~\ref{fig:framework}, V2X-WAM consists of two major stages. Reliability-Aware Cooperative Scene Encoding constructs a shared representation from temporally aligned ego and infrastructure observations while accounting for communication compression and information reliability. Coupled World--Action Modeling subsequently generates multimodal trajectory proposals, predicts the future traffic world conditioned on the selected action, and uses the modeled consequences to refine the final trajectory.

\subsection{Reliability-Aware Cooperative Scene Encoding}
\label{sec:scene_encoding}

The first stage converts heterogeneous vehicle--infrastructure observations into a unified cooperative representation shared by planning and future-world modeling. The ego and infrastructure branches use the same image encoder and temporal BEV encoder, maintaining a consistent feature space while avoiding repeated sensor encoding.

\subsubsection{Temporal Multimodal Agent Encoding}

Camera observations provide appearance and semantic information, whereas temporal LiDAR BEV observations preserve metric geometry and short-term motion. For each agent, the historical BEV sequence is first augmented with consecutive differences. Defining
\begin{equation}
\Delta\mathbf{B}_{j}^{a}
=
\mathbf{B}_{j}^{a}
-
\mathbf{B}_{j-1}^{a},
\label{eq:bev_difference}
\end{equation}
the temporal encoder processes the channel-wise concatenation of
$\mathbf{B}_{t-H+1:t}^{a}$
and
$\Delta\mathbf{B}_{t-H+2:t}^{a}$.

Beyond these temporal differences, the two most recent BEV observations are explicitly compared to capture local motion correspondence. Let
$\mathbf{U}_{a}^{-}$
and
$\mathbf{U}_{a}^{+}$
denote their encoded features. For BEV location $\mathbf{x}$ and displacement
$\boldsymbol{\delta}$ within a local neighborhood $\Omega$, the correlation response is
\begin{equation}
\mathcal{K}^{a}
(\mathbf{x},\boldsymbol{\delta})
=
\left\langle
\mathcal{N}
\left(
\mathbf{U}_{a}^{+}(\mathbf{x})
\right),
\mathcal{N}
\left(
\mathbf{U}_{a}^{-}
(\mathbf{x}+\boldsymbol{\delta})
\right)
\right\rangle ,
\quad
\boldsymbol{\delta}\in\Omega ,
\label{eq:local_correlation}
\end{equation}
where $\mathcal{N}(\cdot)$ denotes channel-wise $\ell_2$ normalization.

Let
$\Delta\mathbf{U}_{a}
=
\mathbf{U}_{a}^{+}-\mathbf{U}_{a}^{-}$.
The correlation volume is fused with the current feature, the signed temporal difference, and its magnitude:
\begin{equation}
\mathbf{G}^{a}
=
\mathbf{G}_{\mathrm{base}}^{a}
+
f_{\mathrm{mot}}
\left(
[
\mathcal{K}^{a};
\mathbf{U}_{a}^{+};
\Delta\mathbf{U}_{a};
|\Delta\mathbf{U}_{a}|
]
\right),
\label{eq:temporal_feature}
\end{equation}
where $[\cdot\,;\cdot]$ denotes channel-wise concatenation. The resulting feature $\mathbf{G}^{a}$ therefore retains both BEV geometry and recent motion correspondence rather than treating historical observations as independent frames.

The current camera image is processed by a shared image encoder and projected to the spatial resolution of $\mathbf{G}^{a}$. For the ego branch, the projected visual feature and $\mathbf{G}^{e}$ are directly fused to obtain the ego-agent representation $\mathbf{E}$. For the infrastructure branch, the temporal BEV feature remains the primary geometric representation because the communicated feature subsequently supports both occupancy and flow prediction. Let $\mathbf{I}_{\mathrm{sem}}$ denote the multimodally fused infrastructure feature. The Semantic Gating module in Fig.~\ref{fig:framework} is defined as
\begin{equation}
\mathbf{I}
=
(1-\alpha_{\mathrm{sem}})
\mathbf{G}^{i}
+
\alpha_{\mathrm{sem}}
\mathbf{I}_{\mathrm{sem}},
\label{eq:semantic_gate}
\end{equation}
where
$\alpha_{\mathrm{sem}}
=
\sigma(\eta_{\mathrm{sem}})$
is a learned scalar gate. This bounded interpolation allows visual semantics to complement infrastructure-side geometry without replacing its temporally consistent BEV structure.

\subsubsection{Reliability-Aware V2X Communication and Fusion}

Communicating the complete infrastructure representation would introduce unnecessary bandwidth overhead. V2X-WAM instead projects $\mathbf{I}$ into a compact message
$\mathbf{C}^{i}\in\mathbb{R}^{C_m\times H_m\times W_m}$,
which is quantized using channel-wise symmetric INT8 quantization.

Let $q_{\max}=2^{7}-1$. For channel $c$, the quantization scale and reconstructed value are
\begin{equation}
\begin{aligned}
&s_c
=
\frac{
\displaystyle
\max_{h,w}
\left|
C^{i}_{c,h,w}
\right|
}{
q_{\max}
},
\\
&\bar C^{i}_{c,h,w}
=
s_c\,
\operatorname{clip}
\left[
\operatorname{round}
\left(
\frac{
C^{i}_{c,h,w}
}{
s_c
}
\right),
-q_{\max},
q_{\max}
\right].
\end{aligned}
\label{eq:quantization}
\end{equation}
During training, a straight-through estimator is used:
\begin{equation}
\mathbf{C}^{q}
=
\mathbf{C}^{i}
+
\operatorname{sg}
\left(
\bar{\mathbf{C}}^{i}
-
\mathbf{C}^{i}
\right),
\label{eq:ste}
\end{equation}
where $\operatorname{sg}(\cdot)$ denotes stop-gradient. Thus, the forward computation follows the quantized communication process while the message encoder retains an identity gradient. The message is subsequently decoded into
$\mathbf{C}^{d}$
in the ego-side feature space.

Infrastructure information is not assumed to contribute uniformly across space or time. The relative configuration between the ego vehicle and infrastructure is represented by
\begin{equation}
\mathbf{g}_{t}
=
[
\Delta x,
\Delta y,
\sin\Delta\psi,
\cos\Delta\psi,
\Delta t,
\kappa_t
],
\label{eq:pair_geometry}
\end{equation}
where
\begin{equation}
\kappa_t
=
\exp
\left(
-\frac{
|\Delta t|
}{
\tau_c
}
\right)
\label{eq:temporal_confidence}
\end{equation}
measures temporal confidence according to the sensor timestamp difference.

After both agents have been aligned to the current ego frame, a learned spatial reliability map is estimated from the ego temporal feature and decoded infrastructure message:
\begin{equation}
\mathbf{R}_{\mathrm{spa}}
=
\sigma
\left(
f_{\mathrm{rel}}
\left(
[
\mathbf{G}^{e};
\mathbf{C}^{d}
]
\right)
\right).
\label{eq:spatial_reliability}
\end{equation}
Spatial and temporal reliability are combined as
\begin{equation}
\mathbf{R}
=
\kappa_t
\mathbf{R}_{\mathrm{spa}},
\qquad
\widetilde{\mathbf{C}}^{i}
=
\mathbf{R}
\odot
\mathbf{C}^{d},
\label{eq:effective_message}
\end{equation}
where $\odot$ denotes element-wise multiplication. Consequently, spatially unreliable regions are attenuated by
$\mathbf{R}_{\mathrm{spa}}$, while temporal misalignment reduces the overall contribution of the infrastructure message through $\kappa_t$. The complete geometry vector $\mathbf{g}_{t}$ is retained separately as contextual information for subsequent planning.

The drivable-area map is independently encoded into
$\mathbf{F}^{\mathrm{map}}$.
Reliability-Aware Cooperative Fusion then produces the shared scene representation:
\begin{equation}
\mathbf{F}^{c}
=
\mathbf{E}
+
f_{\mathrm{fus}}
\left(
[
\mathbf{E};
\widetilde{\mathbf{C}}^{i};
\mathbf{F}^{\mathrm{map}}
]
\right).
\label{eq:scene_fusion}
\end{equation}

Dense future-world prediction benefits from preserving stronger geometric information. V2X-WAM therefore maintains an additional geometry-oriented feature
\begin{equation}
\mathbf{F}^{w}
=
f_{\mathrm{geo}}
\left(
[
\mathbf{G}^{e};
\widetilde{\mathbf{C}}^{i}
]
\right)
+
\alpha_w
\mathbf{F}^{c},
\label{eq:world_feature}
\end{equation}
where
$\alpha_w=\sigma(\eta_w)$
is a learned residual gate. Thus,
$\mathbf{F}^{c}$ provides the semantic cooperative context for global reasoning, whereas
$\mathbf{F}^{w}$ retains a geometry-dominant representation for future occupancy and flow prediction. In parallel, the high-resolution ego BEV feature, the reliability-weighted infrastructure feature, and an upsampled version of $\mathbf{F}^{w}$ are fused into a high-resolution representation
$\mathbf{F}^{h}$
for spatially detailed world decoding.

\subsubsection{Scene Tokenization and Latent Resampling}

The cooperative spatial feature $\mathbf{F}^{c}$ is converted into scene tokens through a convolutional tokenizer and augmented with two-dimensional coordinate embeddings. Three additional condition tokens represent complementary non-spatial information. The first encodes the historical ego positions together with their first- and second-order temporal differences; the second encodes the current ego state together with the normalized pair geometry $\mathbf{g}_{t}$; and the third embeds the high-level route command $r_t$.

Let
$\mathbf{T}\in\mathbb{R}^{N_t\times d}$
denote the complete sequence of scene and condition tokens. Instead of propagating all dense tokens through the downstream modules, V2X-WAM introduces
$N_z$ learned latent queries
$\mathbf{Z}^{(0)}\in\mathbb{R}^{N_z\times d}$.
At resampling layer $\ell$,
\begin{equation}
\mathbf{Z}^{(\ell+1)}
=
\mathcal{A}_{\ell}
\left(
\mathbf{Z}^{(\ell)},
\mathbf{T},
\mathbf{T}
\right),
\label{eq:scene_resampler}
\end{equation}
where
$\mathcal{A}_{\ell}(\mathbf{Q},\mathbf{K},\mathbf{V})$
denotes a cross-attention block. After $L_r$ layers,
$\mathbf{Z}=\mathbf{Z}^{(L_r)}$
forms a fixed-size representation of the cooperative scene and serves as the common latent context for planning and future-world modeling.

\subsection{Multimodal Action Proposal}
\label{sec:action_proposal}

Driving behavior is inherently multimodal, particularly at intersections or in scenes where several feasible maneuvers coexist. V2X-WAM therefore predicts $K$ trajectory hypotheses in parallel rather than directly regressing a single deterministic plan.

We denote the mean-pooled scene latent by
$\bar{\mathbf{z}}
=
N_z^{-1}
\sum_{n=1}^{N_z}\mathbf{z}_n$.
For mode $k$ and future horizon $\tau$, the initial planning query is constructed as
\begin{equation}
\mathbf{q}_{k,\tau}^{(0)}
=
\mathbf{e}_{k}^{\mathrm{mode}}
+
\mathbf{e}_{\tau}^{\mathrm{time}}
+
\lambda_z
\bar{\mathbf{z}},
\label{eq:planning_query}
\end{equation}
where
$\mathbf{e}_{k}^{\mathrm{mode}}$
and
$\mathbf{e}_{\tau}^{\mathrm{time}}$
are learnable mode and horizon embeddings. The complete query set first undergoes self-attention to capture dependencies among modes and horizons and subsequently attends to the shared scene latents $\mathbf{Z}$ through stacked cross-attention blocks. The resulting planning token is denoted by
$\mathbf{q}_{k,\tau}$.

A constant-velocity trajectory prior is constructed from the current physical ego velocity $\mathbf{v}_t$:
\begin{equation}
\mathbf{p}_{\tau}^{\mathrm{cv}}
=
\tau\Delta t\,
\mathbf{v}_{t}.
\label{eq:cv_prior}
\end{equation}

Let
$\bar{\mathbf{q}}_{k}
=
T^{-1}
\sum_{\tau=1}^{T}
\mathbf{q}_{k,\tau}$
denote the temporally pooled token of mode $k$.
Its progress factor is predicted as
\begin{equation}
\rho_k
=
1+
\lambda_{\rho}
\tanh
\left(
f_{\rho}
(
\bar{\mathbf{q}}_{k}
)
\right).
\label{eq:progress_factor}
\end{equation}

Defining the normalized prediction horizon as
$h_{\tau}=\tau/T$,
the $k$-th trajectory candidate is
\begin{equation}
\hat{\mathbf{p}}_{\tau}^{(k)}
=
\rho_k
\mathbf{p}_{\tau}^{\mathrm{cv}}
+
\mathbf{s}_{\mathrm{res}}
h_{\tau}^{\gamma_p}
\odot
\tanh
\left(
f_{\mathrm{res}}
(
\mathbf{q}_{k,\tau}
)
\right),
\label{eq:trajectory_candidate}
\end{equation}
where
$\mathbf{s}_{\mathrm{res}}$
controls the maximum residual displacement and
$\gamma_p>1$
allows progressively greater maneuver flexibility toward longer prediction horizons. Each candidate is finally constrained to the predefined planning range.

A mode-scoring head produces a logit $\ell_k$ for each candidate. The preliminary trajectory is selected as
\begin{equation}
k^{*}
=
\arg\max_{k}\ell_k,
\qquad
\hat{\mathbf{P}}^{(0)}
=
\hat{\mathbf{P}}^{(k^{*})}.
\label{eq:mode_selection}
\end{equation}

Its prospective action sequence is represented by consecutive waypoint increments:
\begin{equation}
\hat{\mathbf{a}}_{\tau}
=
\hat{\mathbf{p}}_{\tau}^{(0)}
-
\hat{\mathbf{p}}_{\tau-1}^{(0)},
\qquad
\hat{\mathbf{p}}_{0}^{(0)}
=
\mathbf{0}.
\label{eq:action_sequence}
\end{equation}
The selected trajectory is therefore an intermediate action proposal rather than the final planning output. Both
$\hat{\mathbf{P}}^{(0)}$
and
$\hat{\mathbf{A}}
=
[\hat{\mathbf{a}}_{1},\ldots,\hat{\mathbf{a}}_{T}]$
subsequently condition future-world prediction.

\subsection{Action-Conditioned Future World Modeling}
\label{sec:world_modeling}

\begin{figure}[pos=htbp]
    \centering
    \includegraphics[width=0.65\linewidth]{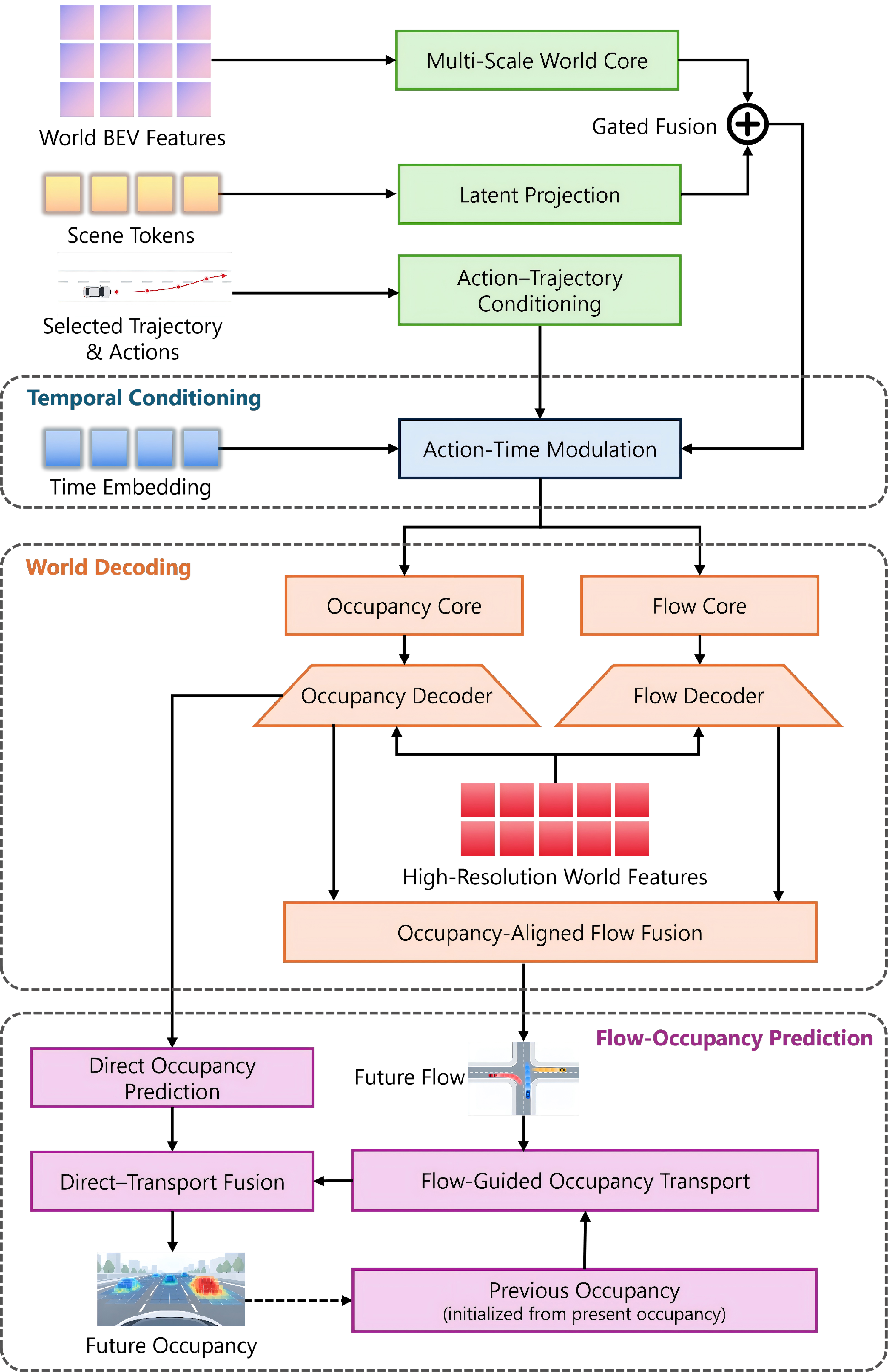}
    \caption{Action-conditioned future-world modeling with joint occupancy--flow prediction and flow-guided occupancy transport.}
    \label{fig:world_decoder}
\end{figure}

The future traffic state depends not only on the current scene but also on the prospective motion of the ego vehicle. V2X-WAM therefore models future occupancy and dynamic flow explicitly conditioned on the selected trajectory and action sequence. The detailed world-modeling process is illustrated in Fig.~\ref{fig:world_decoder}.

\subsubsection{Action--Time Conditioned World State}

The geometry-oriented feature
$\mathbf{F}^{w}$
is first processed by a multi-scale spatial core to enlarge its receptive field. In parallel, the mean-pooled scene latent
$\bar{\mathbf{z}}$
is projected back to the spatial feature dimension. The common world state is constructed as
\begin{equation}
\mathbf{U}
=
f_{\mathrm{ms}}
(
\mathbf{F}^{w}
)
+
\alpha_z
f_z
(
\bar{\mathbf{z}}
),
\label{eq:world_base}
\end{equation}
where
$\alpha_z=\sigma(\eta_z)$,
and the projected latent feature is broadcast over the BEV plane. This representation combines local geometric structure with global cooperative scene context before future-horizon conditioning.

At horizon $\tau$, the preliminary trajectory position and its action increment are normalized by fixed spatial scales
$\mathbf{s}_{p}$
and
$\mathbf{s}_{a}$.
Their joint action representation is
\begin{equation}
\mathbf{e}_{\tau}^{a}
=
f_a
\left(
\left[
\hat{\mathbf{a}}_{\tau}
\oslash
\mathbf{s}_{a};
\hat{\mathbf{p}}_{\tau}^{(0)}
\oslash
\mathbf{s}_{p}
\right]
\right),
\label{eq:action_embedding}
\end{equation}
where $\oslash$ denotes element-wise division.

The action representation is combined with a learnable future-time embedding
$\mathbf{e}_{\tau}^{t}$.
A modulation network then predicts channel-wise scale and shift parameters:
\begin{equation}
[
\boldsymbol{\gamma}_{\tau};
\boldsymbol{\beta}_{\tau}
]
=
f_{\mathrm{mod}}
\left(
\mathbf{e}_{\tau}^{a}
+
\mathbf{e}_{\tau}^{t}
\right).
\label{eq:affine_parameters}
\end{equation}

The common world state is subsequently modulated as
\begin{equation}
\mathbf{U}_{\tau}
=
\mathbf{U}
\odot
\left[
\mathbf{1}
+
\lambda_{\gamma}
\alpha_a
\tanh
(
\boldsymbol{\gamma}_{\tau}
)
\right]
+
\alpha_a
\boldsymbol{\beta}_{\tau},
\label{eq:action_time_modulation}
\end{equation}
where
$\alpha_a=\sigma(\eta_a)$
controls the overall conditioning strength and
$\lambda_{\gamma}$
bounds the multiplicative modulation. This formulation preserves a common spatial world representation while allowing each future horizon to evolve according to both the prospective ego action and elapsed future time.

\subsubsection{Joint Occupancy--Flow Decoding}

For each future horizon, the conditioned state
$\mathbf{U}_{\tau}$
is processed by separate multi-scale occupancy and flow cores. Both branches are subsequently upsampled and fused with the retained high-resolution world feature
$\mathbf{F}^{h}$,
producing decoded features
$\mathbf{H}_{\tau}^{o}$
and
$\mathbf{H}_{\tau}^{v}$.
The occupancy branch directly predicts the future-occupancy logit
$\mathbf{L}_{\tau}^{\mathrm{dir}}$.

Dynamic flow should be estimated at the spatial support of future actors rather than solely from features anchored to their present locations. The decoded flow representation therefore receives a bounded residual from the future occupancy feature:
\begin{equation}
\widetilde{\mathbf{H}}_{\tau}^{v}
=
\mathbf{H}_{\tau}^{v}
+
\alpha_{ov}
\operatorname{sg}
\left(
\mathbf{H}_{\tau}^{o}
\right),
\label{eq:occupancy_flow_fusion}
\end{equation}
where
$\alpha_{ov}
=
\sigma(\eta_{ov})$.
The stop-gradient operator allows the flow branch to exploit future occupancy support without propagating flow supervision back into the occupancy decoder.

The flow head jointly predicts a raw displacement field
$\mathbf{V}_{\tau}^{r}$
and a motion-confidence map
$\mathbf{m}_{\tau}$:
\begin{equation}
\mathbf{V}_{\tau}^{r}
=
v_{\max}
\tanh
\left(
\frac{
f_v
(
\widetilde{\mathbf{H}}_{\tau}^{v}
)
}{
v_{\max}
}
\right),
\label{eq:raw_flow}
\end{equation}
and
\begin{equation}
\mathbf{m}_{\tau}
=
\sigma
\left(
f_m
(
\widetilde{\mathbf{H}}_{\tau}^{v}
)
\right).
\label{eq:motion_confidence}
\end{equation}

The motion confidence is used to suppress unsupported residual motion in static or slowly moving regions. Let
\begin{equation}
\mathbf{d}_{\tau}
=
d_0
+
d_{\max}
(
1-\mathbf{m}_{\tau}
)
\label{eq:adaptive_deadzone}
\end{equation}
denote the adaptive radial dead-zone. The forward displacement is then
\begin{equation}
\mathbf{V}_{\tau}^{s}
=
\left(
1-
\frac{
\mathbf{d}_{\tau}
}{
\|
\mathbf{V}_{\tau}^{r}
\|_2+\epsilon
}
\right)_{+}
\mathbf{V}_{\tau}^{r},
\label{eq:flow_shrinkage}
\end{equation}
where $(x)_{+}=\max(x,0)$. To retain the more stable raw-flow gradient while using the shrunk displacement in the forward pass, the final flow is written as
\begin{equation}
\hat{\mathbf{V}}_{\tau}
=
\mathbf{V}_{\tau}^{r}
+
\operatorname{sg}
\left(
\mathbf{V}_{\tau}^{s}
-
\mathbf{V}_{\tau}^{r}
\right).
\label{eq:flow_ste}
\end{equation}

\subsubsection{Flow-Guided Occupancy Transport}

Direct occupancy prediction provides a flexible future estimate at every horizon, while dynamic flow explicitly describes how occupied regions propagate between consecutive future states. V2X-WAM combines these complementary cues through differentiable occupancy transport.

The directly decoded occupancy probability is
\begin{equation}
\hat{\mathbf{O}}_{\tau}^{\mathrm{dir}}
=
\sigma
\left(
\mathbf{L}_{\tau}^{\mathrm{dir}}
\right).
\label{eq:direct_occupancy}
\end{equation}

Let
$\mathcal{W}(\cdot,\cdot)$
denote differentiable BEV warping implemented by bilinear sampling, where metric flow displacement is converted to the corresponding normalized BEV sampling coordinates. The transported occupancy is
\begin{equation}
\hat{\mathbf{O}}_{\tau}^{\mathrm{tr}}
=
\mathcal{W}
\left(
\operatorname{sg}
(
\hat{\mathbf{O}}_{\tau-1}
),
\hat{\mathbf{V}}_{\tau}
\right).
\label{eq:flow_transport}
\end{equation}
For the first future step, the previous occupancy is initialized from a present-occupancy prediction decoded from the current world state.

The direct and transported predictions are combined using a horizon-dependent learned coefficient:
\begin{equation}
\hat{\mathbf{O}}_{\tau}
=
\mu_{\tau}
\hat{\mathbf{O}}_{\tau}^{\mathrm{dir}}
+
(1-\mu_{\tau})
\hat{\mathbf{O}}_{\tau}^{\mathrm{tr}},
\label{eq:occupancy_fusion}
\end{equation}
where
$\mu_{\tau}
=
\sigma(\eta_{\tau}^{\mathrm{mix}})$.
The direct pathway accommodates newly appearing or uncertain occupancy patterns, whereas the transport pathway promotes temporal consistency according to the predicted motion field.

In addition to dense occupancy and flow, each future horizon produces a compact representation of its predicted consequence. Specifically,
\begin{equation}
\mathbf{w}_{\tau}
=
f_w
\left(
\frac{1}{2}
\operatorname{GAP}
\left(
\mathbf{H}_{\tau}^{o}
+
\mathbf{H}_{\tau}^{v}
\right)
\right),
\label{eq:world_token}
\end{equation}
where
$\operatorname{GAP}(\cdot)$
denotes global average pooling.
The resulting sequence
$\mathbf{W}
=
[\mathbf{w}_{1},\ldots,\mathbf{w}_{T}]$
summarizes the future-world consequences associated with the preliminary action.

\subsection{World-Aware Trajectory Refinement and Joint Optimization}
\label{sec:refinement}

\subsubsection{World-Aware Trajectory Refinement}

The action-conditioned world decoder realizes the action-to-world direction of V2X-WAM. The reverse interaction is established by returning the modeled future consequences to the planning branch.

For each horizon, the selected planning token
$\mathbf{q}_{k^{*},\tau}$
is paired with the corresponding world-consequence token
$\mathbf{w}_{\tau}$.
Defining again
$h_{\tau}=\tau/T$,
the refinement module predicts a bounded trajectory correction:
\begin{equation}
\Delta\hat{\mathbf{p}}_{\tau}
=
\mathbf{s}_{\mathrm{ref}}
h_{\tau}^{\gamma_r}
\odot
\tanh
\left(
f_{\mathrm{ref}}
\left(
[
\mathbf{q}_{k^{*},\tau};
\mathbf{w}_{\tau}
]
\right)
\right),
\label{eq:trajectory_correction}
\end{equation}
where
$\mathbf{s}_{\mathrm{ref}}$
sets the maximum correction in the two planar directions and
$\gamma_r>1$
allows progressively greater refinement at longer horizons.

The final trajectory is obtained as
\begin{equation}
\hat{\mathbf{p}}_{\tau}
=
\operatorname{clip}
\left(
\hat{\mathbf{p}}_{\tau}^{(0)}
+
\Delta\hat{\mathbf{p}}_{\tau},
-\mathbf{s}_{p},
\mathbf{s}_{p}
\right),
\label{eq:trajectory_refinement}
\end{equation}
where
$\mathbf{s}_{p}$
defines the admissible planning range. The bounded refinement preserves the overall maneuver proposed by the multimodal planner while allowing the predicted future consequences to adjust its detailed trajectory. The final action sequence is recomputed from consecutive refined waypoints.

\subsubsection{Cross-Task Gradient Routing}

Planning and dense future-world prediction are explicitly coupled in the forward pass, but their supervision signals differ substantially in density and magnitude. Dense occupancy and flow losses can otherwise dominate the gradient propagated through the shared scene and action interfaces. To regulate this interaction without modifying the forward computation, V2X-WAM employs the gradient-routing operator
\begin{equation}
\mathcal{G}_{\alpha}
(
\mathbf{x}
)
=
\operatorname{sg}
(
\mathbf{x}
)
+
\alpha
\left[
\mathbf{x}
-
\operatorname{sg}
(
\mathbf{x}
)
\right].
\label{eq:gradient_routing}
\end{equation}
The forward value of
$\mathcal{G}_{\alpha}(\mathbf{x})$
is exactly $\mathbf{x}$, while its backward gradient is multiplied by $\alpha$. Independent coefficients are used for the scene-to-planner, action-to-world, and world-to-action connections. The architectural coupling is therefore fully preserved during inference while cross-task gradient interference is controlled during joint optimization.

\subsubsection{Joint Learning Objective}

V2X-WAM is trained in a single stage under joint planning, safety, and future-world supervision. To maintain a compact objective formulation, elementary regression, classification, and regularization terms are grouped according to their functional roles rather than treated as separate task-level losses.

For multimodal planning, the candidate trajectory most consistent with the demonstrated trajectory is first identified. Let
$\mathcal{A}$ denote the set of evaluation-aligned anchor horizons and
$w_j^{a}$ the corresponding horizon weights. The matching cost for candidate
$\hat{\mathbf{P}}^{(k)}$ is defined as
\begin{equation}
\mathcal{C}
\left(
\hat{\mathbf{P}}^{(k)},
\mathbf{P}^{\star}
\right)
=
\frac{
\displaystyle
\sum_{j\in\mathcal{A}}
w_j^{a}
\,
\ell_{\mathrm{SL1}}
\left(
\hat{\mathbf{p}}_{j}^{(k)},
\mathbf{p}_{j}^{\star}
\right)
}{
\displaystyle
\sum_{j\in\mathcal{A}}w_j^{a}
}
+
\lambda_{e}
\left\|
\hat{\mathbf{p}}_{j_{\max}}^{(k)}
-
\mathbf{p}_{j_{\max}}^{\star}
\right\|_2 ,
\label{eq:matching_cost}
\end{equation}
where
$\mathbf{P}^{\star}$
denotes the ground-truth trajectory and
$j_{\max}$ is the longest anchor horizon. The best-matching mode is then
\begin{equation}
k^{\dagger}
=
\arg\min_{k}
\mathcal{C}
\left(
\hat{\mathbf{P}}^{(k)},
\mathbf{P}^{\star}
\right).
\label{eq:best_mode}
\end{equation}

To supervise both dense trajectory evolution and the principal planning horizons, we define a trajectory regression operator
\begin{equation}
\begin{aligned}
\Phi_{\mathrm{traj}}
\left(
\hat{\mathbf{P}},
\mathbf{P}^{\star}
\right)
={}&
\frac{1}{T}
\sum_{\tau=1}^{T}
w_{\tau}^{d}
\,
\ell_{\mathrm{SL1}}
\left(
\hat{\mathbf{p}}_{\tau},
\mathbf{p}_{\tau}^{\star}
\right) +
\frac{
\displaystyle
\sum_{j\in\mathcal{A}}
w_j^{a}
\left\|
\hat{\mathbf{p}}_{j}
-
\mathbf{p}_{j}^{\star}
\right\|_2
}{
\displaystyle
\sum_{j\in\mathcal{A}}w_j^{a}
},
\end{aligned}
\label{eq:trajectory_regression}
\end{equation}
where
$w_{\tau}^{d}$
controls dense temporal supervision.

The multimodal planning objective is subsequently written as
\begin{equation}
\begin{aligned}
\mathcal{L}_{\mathrm{plan}}
={}&
\lambda_{\mathrm{best}}
\Phi_{\mathrm{traj}}
\left(
\hat{\mathbf{P}}^{(k^{\dagger})},
\mathbf{P}^{\star}
\right)
+
\lambda_{\mathrm{top}}
\Phi_{\mathrm{traj}}
\left(
\hat{\mathbf{P}}^{(k^{*})},
\mathbf{P}^{\star}
\right)
+
\lambda_{\mathrm{mode}}
\operatorname{CE}
\left(
\boldsymbol{\ell},
k^{\dagger}
\right)
+
\lambda_{\mathrm{act}}
\operatorname{SL1}
\left(
\hat{\mathbf{A}},
\mathbf{A}^{\star}
\right)
+
\mathcal{R}_{\mathrm{plan}},
\end{aligned}
\label{eq:planning_loss}
\end{equation}
where
$k^{*}$
is the mode selected by Eq.~\eqref{eq:mode_selection},
$\boldsymbol{\ell}$ denotes the mode logits, and
$\mathbf{A}^{\star}$
contains the ground-truth waypoint increments.

The planning regularizer consolidates long-horizon robustness, motion consistency, trajectory smoothness, and mode diversity:
\begin{equation}
\begin{aligned}
\mathcal{R}_{\mathrm{plan}}
={}&
\lambda_{\mathrm{tail}}
\operatorname{MeanTop}_{q}
\left(
\left\|
\hat{\mathbf{p}}_{j_{\max}}^{(k^{*})}
-
\mathbf{p}_{j_{\max}}^{\star}
\right\|_2
\right)
+
\lambda_{\mathrm{prog}}
\left[
\operatorname{SL1}
\left(
\|\Delta\hat{\mathbf{P}}\|_2,
\|\Delta\mathbf{P}^{\star}\|_2
\right)
+
\frac{1}{2}
\operatorname{SL1}
\left(
\hat{\mathbf{D}},
\mathbf{D}^{\star}
\right)
\right]
\\
&+
\lambda_{\mathrm{smooth}}
\left(
\operatorname{Mean}
\left[
\left\|
\Delta^{2}\hat{\mathbf{P}}
\right\|_1
\right]
+
\frac{1}{4}
\operatorname{Mean}
\left[
\left\|
\Delta^{3}\hat{\mathbf{P}}
\right\|_1
\right]
\right)
+
\lambda_{\mathrm{div}}
\operatorname{Mean}_{m\neq n}
\left[
m_{\mathrm{div}}
-
\left\|
\hat{\mathbf{p}}_{j_{\max}}^{(m)}
-
\hat{\mathbf{p}}_{j_{\max}}^{(n)}
\right\|_2
\right]_{+}.
\end{aligned}
\label{eq:planning_regularization}
\end{equation}
Here,
$\Delta\hat{\mathbf{P}}$
and
$\Delta\mathbf{P}^{\star}$
denote consecutive waypoint increments,
$\hat{\mathbf{D}}$
and
$\mathbf{D}^{\star}$
are their cumulative traveled distances,
$\operatorname{MeanTop}_{q}(\cdot)$
averages the largest fraction $q$ of long-horizon errors in a mini-batch, and
$m_{\mathrm{div}}$
is the minimum endpoint-separation margin between trajectory modes.

Safety supervision is applied directly to the refined trajectory. For collision avoidance, let
$r_{b,j}(\mathbf{P})$
denote the differentiable oriented-box collision risk of trajectory
$\mathbf{P}$
for sample $b$ at anchor horizon $j$.
The per-sample collision penalty is
\begin{equation}
c_b
=
\frac{
\displaystyle
\sum_{j\in\mathcal{A}}
w_j^{a}
\left[
\left(
r_{b,j}(\hat{\mathbf{P}})
-
r_{b,j}(\mathbf{P}^{\star})
\right)_{+}
+
\lambda_{\mathrm{abs}}
r_{b,j}(\hat{\mathbf{P}})
\mathbb{I}
\left(
r_{b,j}(\mathbf{P}^{\star})<\delta_c
\right)
\right]
}{
\displaystyle
\sum_{j\in\mathcal{A}}w_j^{a}
}.
\label{eq:collision_penalty}
\end{equation}
This formulation primarily penalizes collision risk introduced beyond that already present in the demonstrated trajectory, while retaining a weaker absolute penalty for otherwise safe ground-truth states.

For road compliance, three laterally displaced footprint points are sampled along each predicted waypoint. Let
$d_{b,\tau,s}(\mathbf{P})$
denote the corresponding off-road cost obtained from the drivable-area map, where
$s$ indexes the sampled footprint points. The complete safety objective is
\begin{equation}
\begin{aligned}
\mathcal{L}_{\mathrm{safe}}
={}&
\lambda_{\mathrm{col}}
\left[
\frac{1}{2B}
\sum_{b=1}^{B}c_b
+
\frac{1}{2}
\operatorname{MeanTop}_{q_c}
\left(
\{c_b\}_{b=1}^{B}
\right)
\right]
+
\lambda_{\mathrm{road}}
\frac{1}{BTS}
\sum_{b,\tau,s}
\Big[
\left(
d_{b,\tau,s}(\hat{\mathbf{P}})
-
d_{b,\tau,s}(\mathbf{P}^{\star})
\right)_{+}
\\
&+
\lambda_{\mathrm{abs}}^{r}
d_{b,\tau,s}(\hat{\mathbf{P}})
\mathbb{I}
\left(
d_{b,\tau,s}(\mathbf{P}^{\star})<\delta_r
\right)
\Big].
\end{aligned}
\label{eq:safety_loss}
\end{equation}

Future-world supervision consists of occupancy and dynamic-flow objectives. Because the final future occupancy, the direct occupancy branch, and the present-occupancy initialization share the same prediction semantics, we first define a common occupancy criterion
\begin{equation}
\Phi_{\mathrm{occ}}
\left(
\hat{\mathbf{O}},
\mathbf{O}^{\star}
\right)
=
\operatorname{HBCE}
\left(
\hat{\mathbf{O}},
\mathbf{O}^{\star}
\right)
+
\lambda_{\mathrm{Tv}}
\operatorname{Tv}
\left(
\hat{\mathbf{O}},
\mathbf{O}^{\star}
\right),
\label{eq:occupancy_criterion}
\end{equation}
where
$\operatorname{HBCE}(\cdot)$
denotes hard-pixel balanced binary cross-entropy. The Tversky term is
\begin{equation}
\operatorname{Tv}
\left(
\hat{\mathbf{O}},
\mathbf{O}^{\star}
\right)
=
1-
\frac{
\mathrm{TP}+\epsilon
}{
\mathrm{TP}
+
\alpha_{\mathrm{Tv}}\mathrm{FP}
+
\beta_{\mathrm{Tv}}\mathrm{FN}
+
\epsilon
},
\label{eq:tversky}
\end{equation}
with
$\mathrm{TP}$,
$\mathrm{FP}$,
and
$\mathrm{FN}$
computed over the valid BEV occupancy cells.

The complete occupancy objective becomes
\begin{equation}
\begin{aligned}
\mathcal{L}_{\mathrm{occ}}
={}&
\Phi_{\mathrm{occ}}
\left(
\hat{\mathbf{O}},
\mathbf{O}^{\star}
\right)
+
\lambda_{\mathrm{dir}}
\Phi_{\mathrm{occ}}
\left(
\hat{\mathbf{O}}^{\mathrm{dir}},
\mathbf{O}^{\star}
\right)
+
\lambda_{\mathrm{pres}}
\Phi_{\mathrm{occ}}
\left(
\hat{\mathbf{O}}^{\mathrm{pres}},
\mathbf{O}^{\star,\mathrm{pres}}
\right),
\end{aligned}
\label{eq:occupancy_loss}
\end{equation}
where
$\hat{\mathbf{O}}^{\mathrm{dir}}$
is the directly decoded future occupancy and
$\hat{\mathbf{O}}^{\mathrm{pres}}$
is the present-occupancy estimate used to initialize the transport process.

Dynamic flow is supervised over valid matched-actor regions. Let
$\Gamma_{\tau}(\mathbf{x})\in\{0,1\}$
denote the foreground flow-validity mask and
$\omega_{\tau}$
the temporal supervision weight.
The local regression error at BEV position
$\mathbf{x}$
is
\begin{equation}
e_{\mathrm{flow}}
(\tau,\mathbf{x})
=
\sum_{d\in\{x,y\}}
\ell_{\mathrm{SL1}}
\left(
\hat V_{\tau,d}(\mathbf{x}),
V_{\tau,d}^{\star}(\mathbf{x})
\right)
+
\lambda_{\mathrm{EPE}}
\left\|
\hat{\mathbf{V}}_{\tau}(\mathbf{x})
-
\mathbf{V}_{\tau}^{\star}(\mathbf{x})
\right\|_2 .
\label{eq:flow_local_error}
\end{equation}

The predicted motion-confidence logit is denoted by
$z_{\tau}(\mathbf{x})$.
Its soft target is derived from the magnitude of the ground-truth displacement:
\begin{equation}
m_{\tau}^{\star}(\mathbf{x})
=
\operatorname{clip}
\left(
\frac{
\|\mathbf{V}_{\tau}^{\star}(\mathbf{x})\|_2-\delta_m
}{
s_m
},
0,1
\right).
\label{eq:motion_target}
\end{equation}

The complete flow objective is then
\begin{equation}
\begin{aligned}
\mathcal{L}_{\mathrm{flow}}
={}&
\frac{
\displaystyle
\sum_{\tau,\mathbf{x}}
\omega_{\tau}
\Gamma_{\tau}(\mathbf{x})
e_{\mathrm{flow}}(\tau,\mathbf{x})
}{
\displaystyle
\sum_{\tau,\mathbf{x}}
\omega_{\tau}
\Gamma_{\tau}(\mathbf{x})
+
\epsilon
}
+
\lambda_{\mathrm{mot}}
\Bigg[
\frac{
\displaystyle
\sum_{\tau,\mathbf{x}}
\omega_{\tau}
\Gamma_{\tau}(\mathbf{x})
\operatorname{BCE}
\left(
z_{\tau}(\mathbf{x}),
m_{\tau}^{\star}(\mathbf{x})
\right)
}{
\displaystyle
\sum_{\tau,\mathbf{x}}
\omega_{\tau}
\Gamma_{\tau}(\mathbf{x})
+
\epsilon
}
\\
&
+
\lambda_{\mathrm{bg}}
\frac{
\displaystyle
\sum_{\tau,\mathbf{x}}
\bar{\Gamma}_{\tau}(\mathbf{x})
\operatorname{BCE}
\left(
z_{\tau}(\mathbf{x}),
0
\right)
}{
\displaystyle
\sum_{\tau,\mathbf{x}}
\bar{\Gamma}_{\tau}(\mathbf{x})
+
\epsilon
}
\Bigg]
+
\lambda_{\mathrm{bg}}
\frac{
\displaystyle
\sum_{\tau,\mathbf{x}}
\bar{\Gamma}_{\tau}(\mathbf{x})
\left\|
\hat{\mathbf{V}}_{\tau}(\mathbf{x})
\right\|_2
}{
\displaystyle
\sum_{\tau,\mathbf{x}}
\bar{\Gamma}_{\tau}(\mathbf{x})
+
\epsilon
}.
\end{aligned}
\label{eq:flow_loss}
\end{equation}
where
$\bar{\Gamma}_{\tau}=1-\Gamma_{\tau}$
denotes unsupported background regions. Motion-confidence supervision encourages the learned motion gate to distinguish genuinely moving actors from static or near-static regions, with background confidence explicitly regularized toward zero. The final background term further suppresses unsupported displacement that could otherwise introduce spurious occupancy during flow-guided transport.

The future-world objective combines occupancy and flow supervision:
\begin{equation}
\mathcal{L}_{\mathrm{world}}
=
\mathcal{L}_{\mathrm{occ}}
+
\lambda_{\mathrm{flow}}
\mathcal{L}_{\mathrm{flow}}.
\label{eq:world_loss}
\end{equation}

Finally, the complete training objective is
\begin{equation}
\mathcal{L}
=
\mathcal{L}_{\mathrm{plan}}
+
\mathcal{L}_{\mathrm{safe}}
+
\lambda_{\mathrm{world}}
\mathcal{L}_{\mathrm{world}}.
\label{eq:total_loss}
\end{equation}
Together with the cross-task gradient routing defined above, this objective jointly optimizes multimodal action generation, safety-aware trajectory refinement, and action-conditioned future-world prediction while controlling interference among their heterogeneous supervision signals.

\section{Experiments}
\label{sec:experiments}

\subsection{Experimental Setup}
\label{sec:exp_setup}

\subsubsection{Dataset and Data Preparation}

We conduct all experiments on V2X-Seq-SPD \citep{yu2023v2xseq}, the sequential perception subset of V2X-Seq. The dataset contains more than 15,000 frames collected from 95 real-world traffic scenes and provides synchronized vehicle- and infrastructure-side images, point clouds, calibration parameters, and sequential object annotations at 10 Hz. We follow the released cooperative train/validation split and report quantitative results on the validation split.

Each sample is constructed from a paired ego-vehicle and infrastructure frame. Historical observations from both agents are transformed into the coordinate frame of the current ego vehicle before being processed by the model. The BEV region covers
$[-50,50]$ m and $[-50,100]$ m along the two planar axes and
$[-4,4]$ m vertically, and is discretized into a
$128\times128$ grid. Each LiDAR BEV frame contains three channels representing occupancy, normalized maximum height, and log-normalized point density. Camera images are resized to
$224\times224$ pixels. Four historical BEV observations are sampled with a stride of five source frames. Future trajectories, occupancy, and dynamic flow are generated at the same five-frame stride for ten future steps, corresponding to a temporal resolution of 0.5 s and a maximum prediction horizon of 5 s. Following the established end-to-end cooperative driving protocol, the principal results are reported at 1, 2, and 3 s.

Future ego waypoints are obtained from the recorded vehicle poses and transformed into the current ego coordinate system. Future occupancy is constructed from cooperative dynamic-object annotations, while dynamic flow describes the displacement of matched actors between consecutive future states. The ego footprint is excluded from the future-world targets so that occupancy and flow supervision focus on the surrounding traffic environment. A rasterized drivable-area map is additionally provided to the model for planning and road-structure reasoning. The high-level route command is derived from the sequence-level navigation destination and discretized into "follow", "left", "right", and "stop", and no future planning waypoint is used to construct this input.

\subsubsection{Implementation Details}

V2X-WAM uses a feature dimension of 256 with eight attention heads. The shared scene representation is compressed into 96 latent tokens, and the planning branch predicts six parallel trajectory modes. The spatial world representation contains 128 channels. Infrastructure information is compressed into a $12\times64\times64$ message and transmitted using the channel-wise INT8 quantization described in Section~\ref{sec:scene_encoding}. Including the 12 channel-wise quantization scales, each message requires 49,200 bytes, corresponding to 98,400 B/s at the communication frequency of 2 Hz.

The gradient-routing coefficients for the scene-to-planner, action-to-world, and world-to-action interfaces are set to 0.03, 0.05, and 0.05, respectively. The model is trained for at most 40 epochs using AdamW with a batch size of 8. The base learning rate is $2\times10^{-4}$, while parameters specific to future-world modeling use a 1.25 learning-rate multiplier. The corresponding weight decays are $10^{-2}$ and $5\times10^{-4}$, respectively. The learning rate is linearly warmed up over the first 6\% of training steps and subsequently decayed using a cosine schedule. The gradient norm is clipped at 2.0, and an exponential moving average with a decay factor of 0.999 is maintained during training. Model selection is based on a joint validation criterion that accounts for both planning and future-world prediction quality, with early stopping applied after eight consecutive epochs without improvement. The loss weights used for joint optimization are summarized in Table~\ref{tab:loss_weights}.

\begin{table}[t]
\centering
\caption{Loss hyperparameters used for training V2X-WAM.}
\label{tab:loss_weights}
\small
\begin{tabular*}{0.55\linewidth}
{@{\extracolsep{\fill}}llc@{}}
\toprule
Objective & Hyperparameter & Value \\
\midrule
\multirow{8}{*}{Planning}
& $\lambda_{\mathrm{best}}$   & 1.00 \\
& $\lambda_{\mathrm{top}}$    & 1.25 \\
& $\lambda_{\mathrm{mode}}$   & 0.35 \\
& $\lambda_{\mathrm{act}}$    & 0.15 \\
& $\lambda_{\mathrm{tail}}$   & 0.70 \\
& $\lambda_{\mathrm{prog}}$   & 0.30 \\
& $\lambda_{\mathrm{smooth}}$ & 0.04 \\
& $\lambda_{\mathrm{div}}$    & 0.015 \\
\midrule
\multirow{2}{*}{Safety}
& $\lambda_{\mathrm{col}}$     & 4.00 \\
& $\lambda_{\mathrm{road}}$    & 1.25 \\
\midrule
\multirow{3}{*}{Occupancy}
& $\lambda_{\mathrm{Tv}}$      & 0.75 \\
& $\lambda_{\mathrm{dir}}$     & 0.20 \\
& $\lambda_{\mathrm{pres}}$    & 0.50 \\
\midrule
\multirow{3}{*}{Flow}
& $\lambda_{\mathrm{EPE}}$     & 0.25 \\
& $\lambda_{\mathrm{mot}}$     & 0.20 \\
& $\lambda_{\mathrm{bg}}$      & 0.005 \\
\midrule
\multirow{2}{*}{Joint}
& $\lambda_{\mathrm{world}}$   & 1.00 \\
& $\lambda_{\mathrm{flow}}$    & 2.00 \\
\bottomrule
\end{tabular*}
\end{table}

To improve robustness to imperfect observations and communication, training includes mild image augmentation, BEV dropout and noise, together with random infrastructure-message removal. During training, we apply photometric image augmentation with a strength of 0.12, spatial BEV block dropout with a probability of 0.04, temporally coherent Gaussian BEV noise with a standard deviation of 0.01, and infrastructure dropout with a probability of 0.03. All augmentation settings are fixed across experiments. The occupancy decision threshold is fixed at 0.30 for both V2X-WAM and the reproduced future-world baselines. Unless explicitly modified in the ablation study, all architectural and optimization settings remain unchanged.

\subsubsection{Evaluation Metrics}

We evaluate V2X-WAM from three complementary aspects: end-to-end planning, future-world prediction, and V2X communication efficiency.

For planning, L2 Error measures the Euclidean distance between predicted and ground-truth ego waypoints. At each reported horizon, the metric averages the waypoint errors from the current time up to that horizon rather than using only the terminal displacement. Collision Rate evaluates the overlap between the predicted ego footprint and future traffic actors, while excluding collisions that are also present along the demonstrated ground-truth trajectory. All planning metrics are computed following the same evaluation protocol as UniMM-V2X, and are reported at 1, 2, and 3 s together with their arithmetic average.

For future-world prediction, Occupied IoU measures the intersection-over-union between predicted and ground-truth occupied BEV cells at each evaluation horizon. Dynamic Flow EPE measures the Euclidean endpoint error of the predicted two-dimensional flow vectors over valid dynamic-actor cells. World metrics are computed only at horizons with valid cooperative annotations. Occupied IoU is reported in percentage points, whereas Dynamic Flow EPE is reported in meters.

Communication efficiency is measured as the average transmitted payload per second. The reported transmission cost includes the quantized infrastructure message and its channel-wise quantization scales and is multiplied by the 2 Hz communication frequency. We report this quantity in bytes per second (B/s).

\subsubsection{Comparison Methods}

For end-to-end planning, we compare V2X-WAM with representative single-agent and cooperative driving methods under the V2X-Seq-SPD protocol. The single-agent group includes VAD \citep{jiang2023vad}, UniAD \citep{hu2023uniad}, and SparseDrive \citep{sun2025sparsedrive}. Cooperative baselines include a vanilla feature-fusion model, V2VNet \citep{wang2020v2vnet}, CooperNaut \citep{cui2022coopernaut}, UniV2X \citep{yu2025univ2x}, and UniMM-V2X \citep{song2026unimmv2x}. The reported baseline planning results and communication costs follow the unified benchmark protocol used by UniMM-V2X.

For future-world prediction, we compare with three representative BEV forecasting methods: FIERY \citep{hu2021fiery}, PowerBEV \citep{li2023powerbev}, and StreamingFlow \citep{shi2024streamingflow}. Their future-prediction formulations are adapted to the same V2X-Seq-SPD target representation and evaluated using the identical occupancy and dynamic-flow protocol.

\subsection{End-to-End Planning Performance}
\label{sec:planning_results}

Table~\ref{tab:planning_results} compares the planning performance and communication cost of V2X-WAM with existing single-agent and cooperative methods. V2X-WAM achieves L2 errors of 0.53, 0.92, and 1.45 m at 1, 2, and 3 s, respectively, resulting in an average error of 0.97 m. Compared with the best-performing cooperative baseline in Table~\ref{tab:planning_results}, UniMM-V2X, the corresponding errors are reduced by 32.1\%, 43.6\%, and 29.3\%, with an overall reduction of 34.9\%. The improvement is particularly pronounced at 2 s, indicating that the cooperative scene representation and world-aware trajectory refinement remain effective beyond immediate short-horizon motion extrapolation.

\begin{table*}[t]
\centering
\caption{End-to-end planning performance and transmission cost on V2X-Seq-SPD.}
\label{tab:planning_results}
\small
\setlength{\tabcolsep}{4.0pt}
\begin{tabular}{@{}lccccccccc@{}}
\toprule
\multirow{2}{*}{Method}
& \multicolumn{4}{c}{L2 Error (m) $\downarrow$}
& \multicolumn{4}{c}{Collision Rate (\%) $\downarrow$}
& \multirow{2}{*}{Trans. Cost (B/s) $\downarrow$} \\
\cmidrule(lr){2-5}
\cmidrule(lr){6-9}
& 1s & 2s & 3s & Avg.
& 1s & 2s & 3s & Avg.
& \\
\midrule
VAD$^{*}$~\citep{jiang2023vad}
& 1.65 & 2.72 & 3.80 & 2.72
& 0.86 & 1.21 & 1.28 & 1.12
& -- \\
UniAD$^{*}$~\citep{hu2023uniad}
& 1.26 & 2.22 & 3.06 & 2.18
& 0.88 & 1.18 & 1.32 & 1.13
& -- \\
SparseDrive$^{*}$~\citep{sun2025sparsedrive}
& 1.02 & 1.69 & 2.37 & 1.69
& 0.46 & 1.23 & 1.28 & 0.99
& -- \\
\midrule
Vanilla
& 1.36 & 2.29 & 3.32 & 2.32
& 1.03 & 0.88 & 1.32 & 1.08
& $8.19\times10^{7}$ \\
V2VNet~\citep{wang2020v2vnet}
& 1.96 & 2.37 & 3.41 & 2.58
& 0.74 & 0.88 & 1.03 & 0.88
& $8.19\times10^{7}$ \\
CooperNaut~\citep{cui2022coopernaut}
& 2.69 & 4.07 & 5.50 & 4.09
& 1.18 & 1.32 & 1.76 & 1.42
& $8.19\times10^{7}$ \\
UniV2X~\citep{yu2025univ2x}
& 1.45 & 2.19 & 3.04 & 2.23
& 0.15 & 0.15 & 0.44 & 0.25
& $8.09\times10^{5}$ \\
UniMM-V2X~\citep{song2026unimmv2x}
& 0.78 & 1.63 & 2.05 & 1.49
& 0.05 & 0.15 & 0.15 & 0.12
& $9.32\times10^{5}$ \\
\midrule
V2X-WAM (Ours)
& \textbf{0.53} & \textbf{0.92} & \textbf{1.45} & \textbf{0.97}
& \textbf{0.00} & \textbf{0.00} & \textbf{0.04} & \textbf{0.01}
& \textbf{98,400} \\
\bottomrule
\multicolumn{10}{l}{\footnotesize $^{*}$Single-agent methods without V2X fusion.}
\end{tabular}
\end{table*}

The gain in trajectory accuracy is accompanied by a substantial improvement in safety. V2X-WAM obtains collision rates of 0.00\%, 0.00\%, and 0.04\% at 1, 2, and 3 s, reducing the average collision rate from 0.12\% for UniMM-V2X to 0.01\%. Importantly, this improvement does not rely on transmitting a dense intermediate feature tensor. V2X-WAM requires only 98,400 B/s, which is 89.4\% lower than the transmission cost of UniMM-V2X and 87.8\% lower than that of UniV2X. The result demonstrates that the compact reliability-aware message can preserve decision-relevant infrastructure information while substantially reducing communication overhead.

Taken together, the planning results support the central design of V2X-WAM: cooperative information is most useful when it is not only incorporated into the current scene representation, but also used to anticipate the consequences of the resulting action and subsequently refine that action.

\subsection{Future-World Prediction Performance}
\label{sec:world_results}

We next evaluate whether the world-modeling branch captures meaningful future traffic evolution rather than serving merely as an auxiliary feature generator. Table~\ref{tab:world_results} reports future occupied-region IoU and dynamic-flow EPE.

\begin{table*}[t]
\centering
\caption{Future-world prediction performance on V2X-Seq-SPD.}
\label{tab:world_results}
\small
\setlength{\tabcolsep}{5.0pt}
\begin{tabular}{@{}lcccccccc@{}}
\toprule
\multirow{2}{*}{Method}
& \multicolumn{4}{c}{Occupied IoU (\%) $\uparrow$}
& \multicolumn{4}{c}{Dynamic Flow EPE (m) $\downarrow$} \\
\cmidrule(lr){2-5}
\cmidrule(lr){6-9}
& 1s & 2s & 3s & Avg.
& 1s & 2s & 3s & Avg. \\
\midrule
FIERY~\citep{hu2021fiery}
& 21.27 & 17.04 & 15.13 & 17.81
& 1.22 & 1.33 & 1.46 & 1.33 \\
PowerBEV~\citep{li2023powerbev}
& 23.46 & 18.85 & 18.32 & 20.21
& 1.03 & 1.14 & 1.20 & 1.12 \\
StreamingFlow~\citep{shi2024streamingflow}
& 24.49 & 19.79 & 17.54 & 20.61
& \textbf{0.93} & \textbf{1.07} & 1.18 & \textbf{1.06} \\
V2X-WAM (Ours)
& \textbf{27.90} & \textbf{21.69} & \textbf{19.63} & \textbf{23.07}
& 1.01 & 1.09 & \textbf{1.16} & 1.08 \\
\bottomrule
\end{tabular}
\end{table*}

V2X-WAM achieves the highest occupied IoU at every evaluation horizon. Its average IoU reaches 23.07\%, exceeding StreamingFlow, the strongest baseline in average occupancy quality, by 2.46 percentage points. At 1 s and 2 s, the improvements over StreamingFlow are 3.41 and 1.90 percentage points, respectively. At 3 s, V2X-WAM reaches 19.63\%, exceeding the strongest baseline at that horizon, PowerBEV, by 1.31 percentage points. The consistent advantage across horizons indicates that action-conditioned cooperative features remain informative as future uncertainty increases.

For dynamic flow, StreamingFlow achieves the lowest average EPE of 1.06 m, while V2X-WAM obtains a closely comparable 1.08 m. V2X-WAM reaches the lowest error at the longest 3 s horizon, with an EPE of 1.16 m compared with 1.18 m for StreamingFlow. This result is consistent with the role of flow in the proposed framework: dynamic flow is not optimized as an isolated prediction endpoint, but also provides the transport field used to propagate occupancy across future horizons. The combination of clearly stronger occupied-region prediction and competitive flow estimation therefore indicates that the joint occupancy--flow representation captures future spatial evolution effectively for downstream world-aware planning.

\subsection{Ablation Study}
\label{sec:ablation}

We conduct a set of controlled ablation experiments to examine the contributions of cooperative information, temporal modeling, reliability estimation, action conditioning, world--action coupling, and flow-guided occupancy transport. Each variant is retrained under the same optimization and evaluation protocol as the full model. Table~\ref{tab:ablation} reports the average 1--3 s planning and future-world metrics.

\begin{table}[t]
\centering
\caption{Ablation study of V2X-WAM.}
\label{tab:ablation}
\small
\setlength{\tabcolsep}{4.5pt}
\begin{tabular}{@{}lcccc@{}}
\toprule
Method
& Avg. L2 Error (m) $\downarrow$
& Avg. Collision (\%) $\downarrow$
& Avg. Occupied IoU (\%) $\uparrow$
& Avg. Flow EPE (m) $\downarrow$ \\
\midrule
Ego Only
& 1.11 & 0.05 & 19.80 & 1.15 \\
w/o Action Conditioning
& 1.15 & 0.02 & 23.02 & 1.09 \\
w/o Flow Transport
& 1.02 & 0.24 & 22.40 & 1.10 \\
w/o Reliability
& 1.22 & 0.10 & 22.15 & 1.15 \\
w/o Temporal Modeling
& 1.10 & 0.08 & 20.58 & 1.46 \\
w/o World--Action Coupling
& 1.08 & 0.02 & 22.82 & 1.14 \\
\midrule
V2X-WAM (Ours)
& \textbf{0.97}
& \textbf{0.01}
& \textbf{23.07}
& \textbf{1.08} \\
\bottomrule
\end{tabular}
\end{table}

The comparison between the full model and Ego Only first confirms the value of cooperative sensing. Removing infrastructure information increases average L2 error from 0.97 to 1.11 m and reduces occupied IoU by 3.27 percentage points. The simultaneous degradation in planning and future-world prediction indicates that the infrastructure branch contributes information that cannot be fully recovered from ego observations alone.

Reliability estimation has a particularly strong effect on planning. Without the learned reliability mechanism, average L2 error rises to 1.22 m, the largest planning degradation among the tested variants, while collision rate increases from 0.01\% to 0.10\%. Occupied IoU also decreases to 22.15\%. These results show that simply providing infrastructure features is insufficient; their contribution needs to be regulated according to spatial consistency and temporal synchronization.

Temporal modeling is most critical for representing world dynamics. Removing temporal history increases dynamic-flow EPE from 1.08 to 1.46 m and reduces occupied IoU from 23.07\% to 20.58\%. Planning error and collision rate also increase. This pronounced degradation in flow prediction supports the use of temporal BEV histories and local correspondence modeling for recovering short-term motion cues before future-world decoding.

The two directions of world--action interaction provide complementary effects. Removing only Action$\rightarrow$World conditioning increases average L2 error from 0.97 to 1.15 m even though aggregate occupancy and flow metrics change only slightly. This indicates that conditioning the predicted future on the prospective action primarily improves the relevance of the imagined world to the planning process, rather than merely improving unconditional prediction accuracy. When the complete world--action coupling is removed, average L2 error increases to 1.08 m and flow EPE to 1.14 m. The results support explicit interaction between action generation and future-world reasoning rather than treating the two as independent heads of a shared encoder.

Flow-guided occupancy transport exhibits a different behavior. Removing it produces a relatively modest increase in L2 error and a 0.67-point decrease in occupied IoU, but raises the average collision rate from 0.01\% to 0.24\%. This contrast suggests that temporal consistency of predicted occupied regions is especially important for safety-sensitive trajectory refinement, even when aggregate geometric planning error changes only moderately. Overall, the complete V2X-WAM achieves the strongest joint performance across planning, safety, occupancy, and flow metrics.

\subsection{Qualitative Analysis}
\label{sec:qualitative}

\subsubsection{Joint Planning and Future-Occupancy Prediction}

\begin{figure}[pos=htbp]
    \centering
    \includegraphics[width=0.8\linewidth]{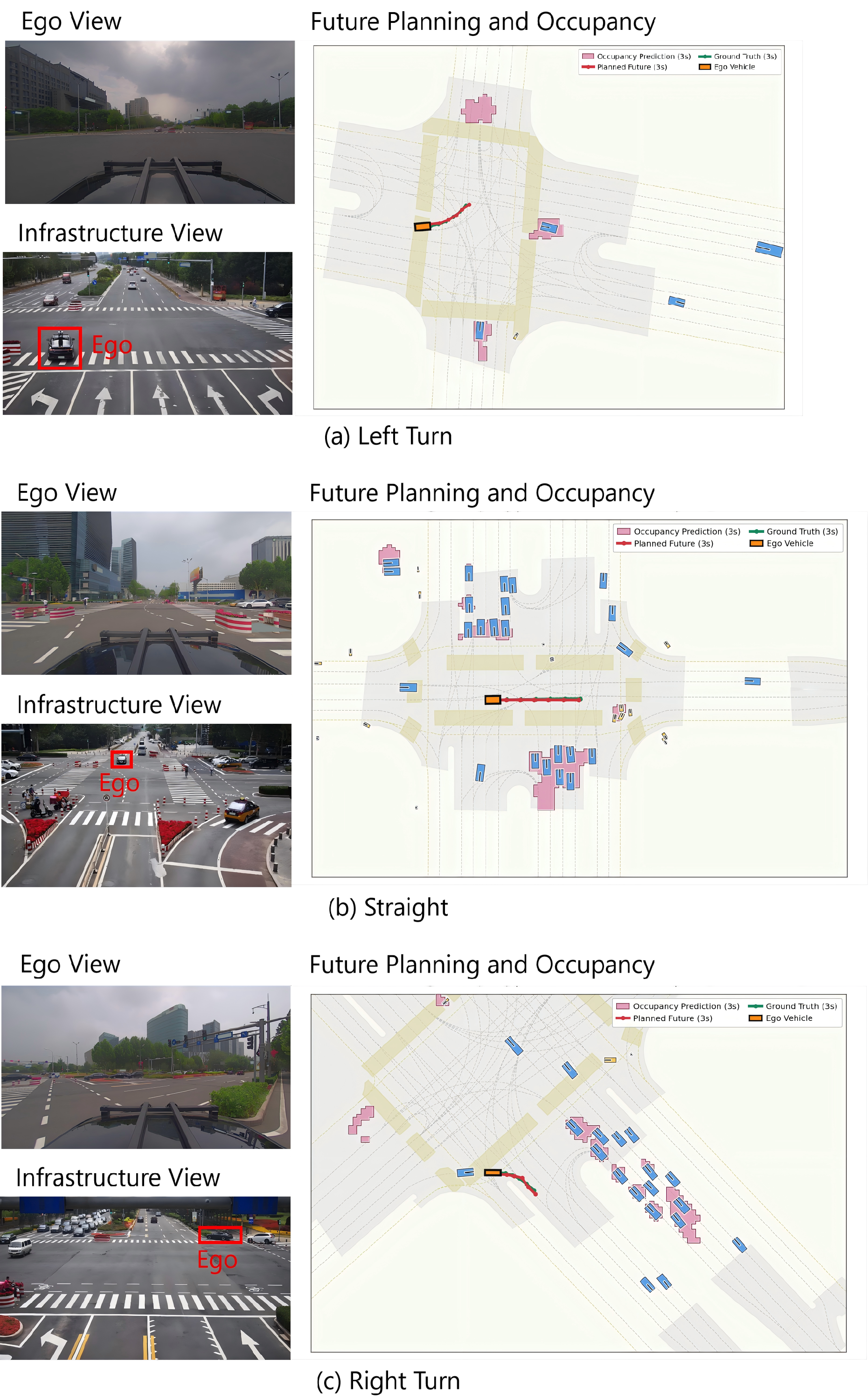}
    \caption{Qualitative planning and 3-s future-occupancy results for representative left-turn, straight, and right-turn scenes.}
    \label{fig:planning_occ}
\end{figure}

Figure~\ref{fig:planning_occ} presents representative qualitative results for left-turn, straight-driving, and right-turn scenarios. For each scene, the ego and infrastructure camera views are shown together with the corresponding BEV visualization. The infrastructure view provides an elevated view of the intersection and complements traffic context that is partially or fully outside the ego camera's immediate field of view. The BEV panel overlays the complete 3 s planned trajectory, the ground-truth trajectory, and the predicted occupancy at 3 s in the same current-ego coordinate frame.

In the left-turn example, the predicted trajectory follows the turning geometry while remaining spatially separated from the forecast occupied regions around the intersection. In the straight-driving scene, the planner maintains a consistent path through the junction while the world branch captures future occupancy associated with surrounding cross-traffic. The right-turn case similarly shows a curved trajectory that remains coherent with both the drivable region and predicted future traffic occupancy. Across the three maneuver types, the planned trajectories remain close to the demonstrated paths while responding to the spatial structure of the predicted future world.

These examples also illustrate the functional connection between the two outputs of V2X-WAM. Future occupancy is not produced solely for visualization or auxiliary supervision; it represents the world consequence associated with the preliminary action and contributes to the subsequent trajectory refinement. The qualitative consistency between planned motion and future occupied regions complements the quantitative planning and world-prediction improvements reported in Tables~\ref{tab:planning_results} and~\ref{tab:world_results}.

\subsubsection{Contribution of Infrastructure Information to Future-World Modeling}

\begin{figure}[pos=htbp]
    \centering
    \includegraphics[width=\textwidth]{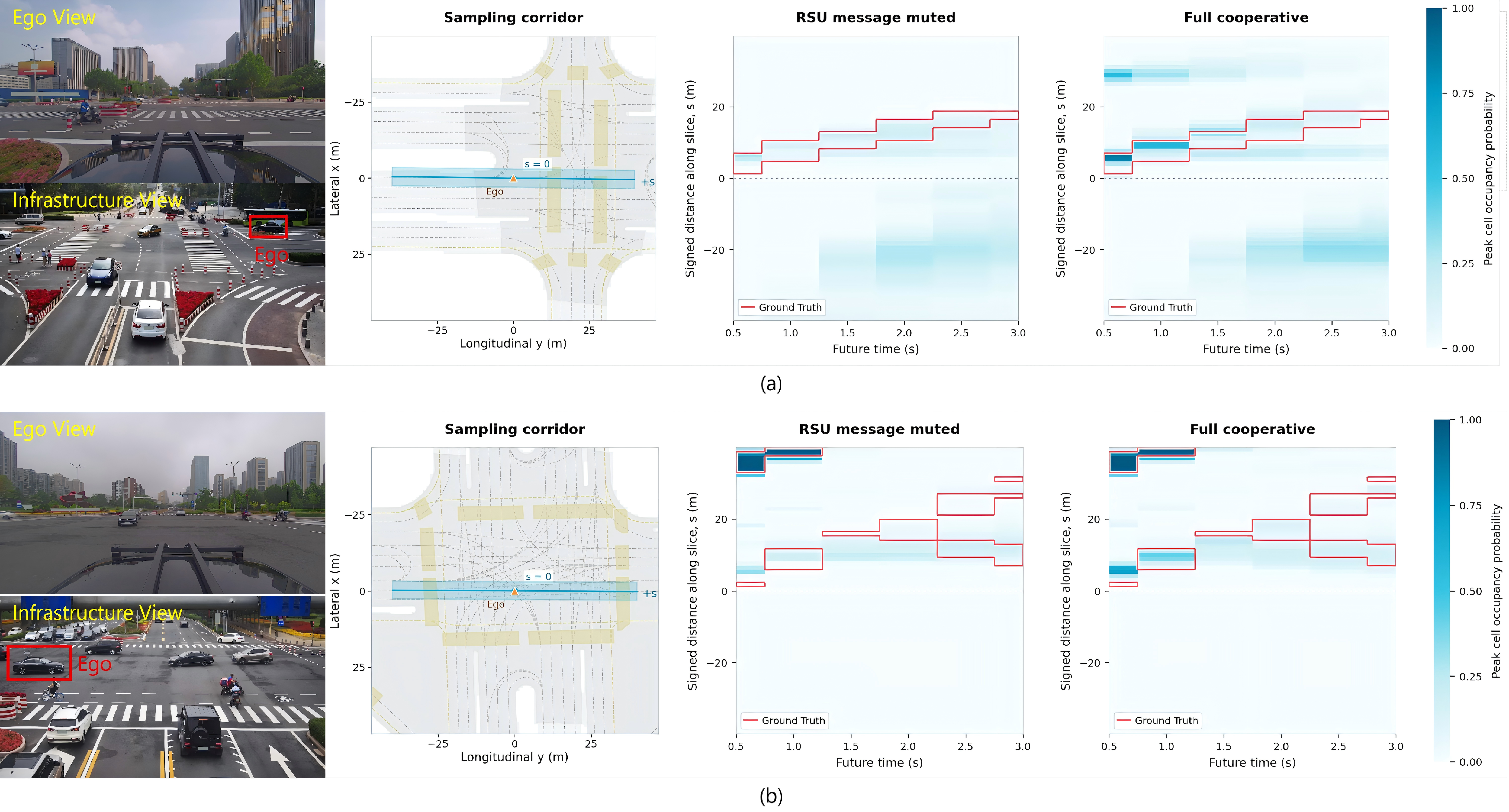}
    \caption{Controlled infrastructure-message analysis of future occupancy along road-aligned space--time slices.}
    \label{fig:spacetime}
\end{figure}

To examine how the infrastructure message contributes specifically to future-world prediction, we perform a controlled feature-level analysis shown in Fig.~\ref{fig:spacetime}. For each scene, V2X-WAM is first evaluated with full cooperative information. The preliminary planner representation and action sequence are then held fixed while only the decoded roadside-unit message entering the world-modeling branch is muted. Consequently, the two conditions differ in infrastructure information available to the future-world model while sharing the same prospective ego action.

For visualization, a straight sampling corridor is constructed from the nearest current HD-map lane tangent. The corridor is 6 m wide and extends 40 m in both directions from its anchor point. Native BEV cells within this corridor are projected onto signed longitudinal-distance bins, producing a space--time view over the 0.5--3.0 s prediction horizon. Color intensity represents the maximum predicted occupancy score within each longitudinal bin, while the red outlines indicate ground-truth occupied regions. Neither future predictions nor ground-truth occupancy are used to select the corridor.

The two examples in Fig.~\ref{fig:spacetime} show a clear difference between the muted and fully cooperative conditions. With the roadside message removed, several future occupied regions become weak or spatially diffuse along the road-aligned slice. Restoring the cooperative message produces stronger and more structured occupancy responses that better follow the temporal progression of the annotated traffic regions. The effect is particularly visible at portions of the corridor where the infrastructure view provides a clearer observation of surrounding traffic than the ego view.

This controlled comparison complements the retrained Ego Only experiment in Table~\ref{tab:ablation}. Whereas Ego Only evaluates the performance of a model learned without infrastructure information, Fig.~\ref{fig:spacetime} isolates the immediate contribution of the infrastructure message to future-world reasoning within the same trained V2X-WAM and under an unchanged action proposal. Together, the two analyses show that cooperative information contributes not only to the initial scene representation, but also directly to the model's representation of how the traffic environment may evolve under the planned action.

\section{Conclusion}
\label{sec:conclusion}

This paper presented V2X-WAM, a cooperative world action model for end-to-end autonomous driving. The proposed framework moves beyond directly mapping cooperative observations to driving actions by explicitly modeling the interaction between prospective actions and future traffic evolution. V2X-WAM constructs a reliability-aware cooperative representation from temporal vehicle- and infrastructure-side observations, compresses roadside information into a compact quantized message, and generates multimodal trajectory proposals from the resulting scene representation. The selected prospective action further conditions future occupancy and dynamic-flow prediction, while the predicted world consequences are fed back to refine the planned trajectory. In this way, cooperative perception, planning, and future-world reasoning are integrated within a unified world--action modeling framework.

Experiments on a large-scale real-world cooperative driving dataset demonstrate consistent improvements in planning accuracy and safety over representative single-agent and cooperative end-to-end driving methods, together with substantially reduced communication overhead. V2X-WAM also achieves strong future-world prediction performance, while ablation studies confirm the effectiveness of the major components and their complementary roles. Overall, the results show that the value of vehicle--infrastructure cooperation extends beyond improving current-scene understanding: cooperative information can also support explicit reasoning about how prospective driving actions interact with the evolving traffic environment. Future work will further investigate richer multi-agent interactions, longer-horizon world modeling, and closed-loop evaluation under more diverse and challenging traffic conditions.



\section*{Acknowledgements}

This work was supported in part by the 
National Natural Science Foundation of China under Grant 52502420.

\printcredits

\bibliographystyle{cas-model2-names}

\bibliography{cas-refs}

@inproceedings{shao2024lmdrive,
  title     = {{LMDrive}: Closed-Loop End-to-End Driving with Large Language Models},
  author    = {Shao, Hao and Hu, Yuxuan and Wang, Letian and Song, Guanglu and
               Waslander, Steven L. and Liu, Yu and Li, Hongsheng},
  booktitle = {Proceedings of the IEEE/CVF Conference on Computer Vision and Pattern Recognition},
  pages     = {15120--15130},
  year      = {2024},
  doi       = {10.1109/CVPR52733.2024.01432}
}

@article{hwang2025emma,
  title   = {{EMMA}: End-to-End Multimodal Model for Autonomous Driving},
  author  = {Hwang, Jyh-Jing and Xu, Runsheng and Lin, Hubert and
             Hung, Wei-Chih and Ji, Jingwei and Choi, Kristy and
             Huang, Di and He, Tong and Covington, Paul and
             Sapp, Benjamin and Zhou, Yin and Guo, James and
             Anguelov, Dragomir and Tan, Mingxing},
  journal = {Transactions on Machine Learning Research},
  year    = {2025}
}

@inproceedings{zhou2025autovla,
  title     = {{AutoVLA}: A Vision-Language-Action Model for End-to-End
               Autonomous Driving with Adaptive Reasoning and Reinforcement Fine-Tuning},
  author    = {Zhou, Zewei and Cai, Tianhui and Zhao, Seth Z. and Zhang, Yun and
               Huang, Zhiyu and Zhou, Bolei and Ma, Jiaqi},
  booktitle = {Advances in Neural Information Processing Systems},
  volume    = {38},
  pages     = {31725--31761},
  year      = {2025},
  doi       = {10.52202/085713-0942}
}

@inproceedings{zhou2026opendrivevla,
  title     = {{OpenDriveVLA}: Towards End-to-End Autonomous Driving with
               Large Vision Language Action Model},
  author    = {Zhou, Xingcheng and Han, Xuyuan and Yang, Feng and
               Ma, Yunpu and Tresp, Volker and Knoll, Alois},
  booktitle = {Proceedings of the AAAI Conference on Artificial Intelligence},
  volume    = {40},
  number    = {16},
  pages     = {13782--13790},
  year      = {2026},
  doi       = {10.1609/aaai.v40i16.38386}
}

@inproceedings{wang2024drivewm,
  title     = {Driving into the Future: Multiview Visual Forecasting and
               Planning with World Model for Autonomous Driving},
  author    = {Wang, Yuqi and He, Jiawei and Fan, Lue and Li, Hongxin and
               Chen, Yuntao and Zhang, Zhaoxiang},
  booktitle = {Proceedings of the IEEE/CVF Conference on Computer Vision and Pattern Recognition},
  pages     = {14749--14759},
  year      = {2024}
}

@inproceedings{li2025law,
  title     = {Enhancing End-to-End Autonomous Driving with Latent World Model},
  author    = {Li, Yingyan and Fan, Lue and He, Jiawei and Wang, Yuqi and
               Chen, Yuntao and Zhang, Zhaoxiang and Tan, Tieniu},
  booktitle = {International Conference on Learning Representations},
  year      = {2025}
}

@inproceedings{zheng2025world4drive,
  title     = {{World4Drive}: End-to-End Autonomous Driving via
               Intention-Aware Physical Latent World Model},
  author    = {Zheng, Yupeng and Yang, Pengxuan and Xing, Zebin and
               Zhang, Qichao and Zheng, Yuhang and Gao, Yinfeng and
               Li, Pengfei and Zhang, Teng and Xia, Zhongpu and
               Jia, Peng and Lang, XianPeng and Zhao, Dongbin},
  booktitle = {Proceedings of the IEEE/CVF International Conference on Computer Vision},
  pages     = {28632--28642},
  year      = {2025},
  doi       = {10.1109/ICCV51701.2025.02659}
}

@inproceedings{tan2026ldrive,
  title     = {Latent Chain-of-Thought World Modeling for End-to-End Autonomous Driving},
  author    = {Tan, Shuhan and Chitta, Kashyap and Chen, Yuxiao and
               Tian, Ran and You, Yurong and Wang, Yan and Luo, Wenjie and
               Cao, Yulong and Kr{\"a}henb{\"u}hl, Philipp and
               Pavone, Marco and Ivanovic, Boris},
  booktitle = {Proceedings of the IEEE/CVF Conference on Computer Vision and Pattern Recognition},
  pages     = {39724--39733},
  year      = {2026}
}

@article{wang2026latentwam,
  title         = {{Latent-WAM}: Latent World Action Modeling for
                   End-to-End Autonomous Driving},
  author        = {Wang, Linbo and Zheng, Yupeng and Chen, Qiang and
                   Li, Shiwei and Zhang, Yichen and Xing, Zebin and
                   Zhang, Qichao and Li, Xiang and Qian, Deheng and
                   Yang, Pengxuan and Dong, Yihang and Hao, Ce and
                   Ye, Xiaoqing and Han, Junyu and Pan, Yifeng and
                   Zhao, Dongbin},
  journal       = {arXiv preprint arXiv:2603.24581},
  year          = {2026},
  eprint        = {2603.24581},
  archivePrefix = {arXiv},
  primaryClass  = {cs.CV}
}

@inproceedings{yu2025univ2x,
  title     = {End-to-End Autonomous Driving Through {V2X} Cooperation},
  author    = {Yu, Haibao and Yang, Wenxian and Zhong, Jiaru and
               Yang, Zhenwei and Fan, Siqi and Luo, Ping and Nie, Zaiqing},
  booktitle = {Proceedings of the AAAI Conference on Artificial Intelligence},
  volume    = {39},
  number    = {9},
  pages     = {9598--9606},
  year      = {2025},
  doi       = {10.1609/aaai.v39i9.33040}
}

@inproceedings{song2026unimmv2x,
  title     = {{UniMM-V2X}: MoE-Enhanced Multi-Level Fusion for
               End-to-End Cooperative Autonomous Driving},
  author    = {Song, Ziyi and Xia, Chen and Wang, Chenbing and
               Yu, Haibao and Zhou, Sheng and Niu, Zhisheng},
  booktitle = {Proceedings of the AAAI Conference on Artificial Intelligence},
  volume    = {40},
  number    = {11},
  pages     = {9135--9143},
  year      = {2026},
  doi       = {10.1609/aaai.v40i11.37870}
}

@article{you2026v2xvlm,
  title   = {{V2X-VLM}: End-to-End {V2X} Cooperative Autonomous Driving
             Through Large Vision-Language Models},
  author  = {You, Junwei and Jiang, Zhuoyu and Huang, Zilin and
             Shi, Haotian and Gan, Rui and Wu, Keshu and Cheng, Xi and
             Li, Xiaopeng and Ran, Bin},
  journal = {Transportation Research Part C: Emerging Technologies},
  volume  = {183},
  pages   = {105457},
  year    = {2026},
  doi     = {10.1016/j.trc.2025.105457}
}

@article{peng2026omniv2x,
  title         = {{OmniV2X}: A Generative Foundation Planner for
                   Efficient End-to-End Cooperative Driving},
  author        = {Peng, Juntong and Lu, Juanwu and Zhou, Yupeng and
                   Cui, Can and Chen, Yaobin and Wang, Ziran},
  journal       = {arXiv preprint arXiv:2606.21165},
  year          = {2026},
  eprint        = {2606.21165},
  archivePrefix = {arXiv},
  primaryClass  = {cs.CV}
}

@inproceedings{song2026dhvlm,
  title     = {{DH-VLM}: Dual-Horizon Cooperative Latent Reasoning
               for Autonomous Driving},
  author    = {Song, Ziyi and Xia, Chen and Yu, Hang and
               Zhou, Sheng and Niu, Zhisheng},
  booktitle = {Computer Vision -- ECCV 2026},
  pages     = {206--224},
  year      = {2026},
  publisher = {Springer}
}

@inproceedings{tian2025drivevlm,
  title     = {{DriveVLM}: The Convergence of Autonomous Driving and Large Vision-Language Models},
  author    = {Tian, Xiaoyu and Gu, Junru and Li, Bailin and Liu, Yicheng and
               Wang, Yang and Zhao, Zhiyong and Zhan, Kun and Jia, Peng and
               Lang, XianPeng and Zhao, Hang},
  booktitle = {Proceedings of The 8th Conference on Robot Learning},
  series    = {Proceedings of Machine Learning Research},
  volume    = {270},
  pages     = {4698--4726},
  publisher = {PMLR},
  year      = {2025}
}

@article{jiang2024senna,
  title         = {Senna: Bridging Large Vision-Language Models and End-to-End Autonomous Driving},
  author        = {Jiang, Bo and Chen, Shaoyu and Liao, Bencheng and
                   Zhang, Xingyu and Yin, Wei and Zhang, Qian and
                   Huang, Chang and Liu, Wenyu and Wang, Xinggang},
  journal       = {arXiv preprint arXiv:2410.22313},
  year          = {2024},
  eprint        = {2410.22313},
  archivePrefix = {arXiv},
  primaryClass  = {cs.CV}
}

@inproceedings{wang2025omnidrive,
  title     = {{OmniDrive}: A Holistic Vision-Language Dataset for Autonomous Driving with Counterfactual Reasoning},
  author    = {Wang, Shihao and Yu, Zhiding and Jiang, Xiaohui and
               Lan, Shiyi and Shi, Min and Chang, Nadine and
               Kautz, Jan and Li, Ying and Alvarez, Jose M.},
  booktitle = {Proceedings of the IEEE/CVF Conference on Computer Vision and Pattern Recognition},
  pages     = {22442--22452},
  year      = {2025}
}

@inproceedings{yang2026drivemoe,
  title     = {{DriveMoE}: Mixture-of-Experts for Vision-Language-Action Model
               in End-to-End Autonomous Driving},
  author    = {Yang, Zhenjie and Chai, Yilin and Jia, Xiaosong and
               Li, Qifeng and Shao, Yuqian and Zhu, Xuekai and
               Su, Haisheng and Yan, Junchi},
  booktitle = {Proceedings of the IEEE/CVF Conference on Computer Vision and Pattern Recognition},
  pages     = {10678--10688},
  year      = {2026}
}

@inproceedings{wang2026linkvla,
  title     = {Unifying Language-Action Understanding and Generation for Autonomous Driving},
  author    = {Wang, Xinyang and Liu, Qian and Ding, Wenjie and
               Yang, Zhao and Li, Wei and Liu, Chang and
               Li, Bailin and Zhan, Kun and Lang, Xianpeng and Chen, Wei},
  booktitle = {Proceedings of the IEEE/CVF Conference on Computer Vision and Pattern Recognition},
  pages     = {25193--25203},
  year      = {2026}
}

@inproceedings{xia2026drivelaw,
  title     = {{DriveLaW}: Unifying Planning and Video Generation in a Latent Driving World},
  author    = {Xia, Tianze and Li, Yongkang and Zhou, Lijun and
               Yao, Jingfeng and Xiong, Kaixin and Sun, Haiyang and
               Wang, Bing and Ma, Kun and Chen, Guang and Ye, Hangjun and
               Liu, Wenyu and Wang, Xinggang},
  booktitle = {Proceedings of the IEEE/CVF Conference on Computer Vision and Pattern Recognition},
  pages     = {39701--39712},
  year      = {2026}
}

@article{wang2026vlaworld,
  title         = {Learning Vision-Language-Action World Models for Autonomous Driving},
  author        = {Wang, Guoqing and Tang, Pin and Ren, Xiangxuan and
                   Zhao, Guodongfang and Feng, Bailan and Ma, Chao},
  journal       = {arXiv preprint arXiv:2604.09059},
  year          = {2026},
  eprint        = {2604.09059},
  archivePrefix = {arXiv},
  primaryClass  = {cs.CV}
}

@article{li2026metis,
  title         = {Metis: A Generalizable and Efficient World-Action Model
                   for Autonomous Driving and Urban Navigation},
  author        = {Li, Jingyu and Liu, Zhe and Hu, Dongnan and
                   Wu, Junjie and Ma, Zipei and Wu, Wenxiao and
                   Han, Chao and Hao, Zhihui and Liu, Zhikang and
                   Zhan, Kun and Deng, Jiankang and Zhu, Xiatian and Zhang, Li},
  journal       = {arXiv preprint arXiv:2606.15869},
  year          = {2026},
  eprint        = {2606.15869},
  archivePrefix = {arXiv},
  primaryClass  = {cs.CV}
}

@article{zhao2026simwam,
  title         = {{SimWAM}: A Simple World Action Model for End-to-End Autonomous Driving},
  author        = {Zhao, Zongchuang and Zhou, Xin and Xu, Tianyang and
                   Sun, Zhengyang and Zhou, Kaixuan and Li, Honglin and
                   Liang, Dingkang and Bai, Xiang},
  journal       = {arXiv preprint arXiv:2608.07468},
  year          = {2026},
  eprint        = {2608.07468},
  archivePrefix = {arXiv},
  primaryClass  = {cs.CV}
}

@inproceedings{cui2022coopernaut,
  title     = {{COOPERNAUT}: End-to-End Driving with Cooperative Perception
               for Networked Vehicles},
  author    = {Cui, Jiaxun and Qiu, Hang and Chen, Dian and
               Stone, Peter and Zhu, Yuke},
  booktitle = {Proceedings of the IEEE/CVF Conference on Computer Vision and Pattern Recognition},
  pages     = {17252--17262},
  year      = {2022},
  doi       = {10.1109/CVPR52688.2022.01674}
}

@article{liu2025codriving,
  title   = {Toward Collaborative Autonomous Driving:
             Simulation Platform and End-to-End System},
  author  = {Liu, Genjia and Hu, Yue and Xu, Chenxin and
             Mao, Weibo and Ge, Junhao and Huang, Zhengxiang and
             Lu, Yifan and Xu, Yinda and Xia, Junkai and
             Wang, Yafei and Chen, Siheng},
  journal = {IEEE Transactions on Pattern Analysis and Machine Intelligence},
  volume  = {47},
  number  = {8},
  pages   = {6566--6584},
  year    = {2025},
  doi     = {10.1109/TPAMI.2025.3560327}
}

@inproceedings{luo2026v2xunipool,
  title     = {{V2X-UniPool}: Unifying Multimodal Perception and
               Knowledge Reasoning for Autonomous Driving},
  author    = {Luo, Xuewen and Yang, Fengze and Fan, Ding and
               Gao, Xiangbo and Yu, Bo and Li, Zihao and
               Tu, Zhengzhong and Zhou, Yang and Liu, Chenxi},
  booktitle = {Proceedings of the IEEE/CVF Conference on Computer Vision and Pattern Recognition Workshops},
  pages     = {747--756},
  year      = {2026}
}

@article{xu2026aurora,
  title         = {Roadside-Cooperative Autonomous Driving:
                   From Data Platform to Vision-Language End-to-End Reasoning},
  author        = {Xu, Yitao and Wu, Tong and Wu, Yiyan and Xu, Guoji and
                   Jiang, Yanbo and Wang, Jiahao and Ke, Zehong and
                   Jiang, Junkai and Zhang, Fang and Wang, Jianqiang},
  journal       = {arXiv preprint arXiv:2608.21032},
  year          = {2026},
  eprint        = {2608.21032},
  archivePrefix = {arXiv},
  primaryClass  = {cs.CV}
}

@article{you2026seal,
  title   = {{SEAL}: Vision-Language Model-Based Safe End-to-End Cooperative
             Autonomous Driving with Adaptive Long-Tail Modeling},
  author  = {You, Junwei and Li, Pei and Jiang, Zhuoyu and Huang, Zilin
             and Gan, Rui and Shi, Haotian and Ran, Bin},
  journal = {Accident Analysis \& Prevention},
  volume  = {238},
  pages   = {108748},
  year    = {2026},
  doi     = {10.1016/j.aap.2026.108748}
}

@inproceedings{yu2023v2xseq,
  title     = {V2X-Seq: A Large-Scale Sequential Dataset for Vehicle-Infrastructure Cooperative Perception and Forecasting},
  author    = {Yu, Haibao and Yang, Wenxian and Ruan, Hongzhi and Yang, Zhenwei and Tang, Yingjuan and Gao, Xu and Hao, Xin and Shi, Yifeng and Pan, Yifeng and Sun, Ning and Song, Juan and Yuan, Jirui and Luo, Ping and Nie, Zaiqing},
  booktitle = {Proceedings of the IEEE/CVF Conference on Computer Vision and Pattern Recognition},
  pages     = {5486--5495},
  year      = {2023}
}

@inproceedings{hu2021fiery,
  title     = {FIERY: Future Instance Prediction in Bird's-Eye View from Surround Monocular Cameras},
  author    = {Hu, Anthony and Murez, Zak and Mohan, Nikhil and Dudas, Sof{\'i}a and Hawke, Jeffrey and Badrinarayanan, Vijay and Cipolla, Roberto and Kendall, Alex},
  booktitle = {Proceedings of the IEEE/CVF International Conference on Computer Vision},
  pages     = {15273--15282},
  year      = {2021}
}

@inproceedings{li2023powerbev,
  title     = {PowerBEV: A Powerful Yet Lightweight Framework for Instance Prediction in Bird's-Eye View},
  author    = {Li, Peizheng and Ding, Shuxiao and Chen, Xieyuanli and Hanselmann, Niklas and Cordts, Marius and Gall, Juergen},
  booktitle = {Proceedings of the Thirty-Second International Joint Conference on Artificial Intelligence},
  pages     = {1080--1088},
  year      = {2023},
  doi       = {10.24963/ijcai.2023/120}
}

@inproceedings{shi2024streamingflow,
  title     = {StreamingFlow: Streaming Occupancy Forecasting with Asynchronous Multi-modal Data Streams via Neural Ordinary Differential Equation},
  author    = {Shi, Yining and Jiang, Kun and Wang, Ke and Li, Jiusi and Wang, Yunlong and Yang, Mengmeng and Yang, Diange},
  booktitle = {Proceedings of the IEEE/CVF Conference on Computer Vision and Pattern Recognition},
  pages     = {14833--14842},
  year      = {2024}
}

@inproceedings{sun2025sparsedrive,
  title     = {SparseDrive: End-to-End Autonomous Driving via Sparse Scene Representation},
  author    = {Sun, Wenchao and Lin, Xuewu and Shi, Yining and Zhang, Chuang and Wu, Haoran and Zheng, Sifa},
  booktitle = {2025 IEEE International Conference on Robotics and Automation (ICRA)},
  pages     = {8795--8801},
  year      = {2025},
  organization = {IEEE},
  doi       = {10.1109/ICRA55743.2025.11128800}
}

@inproceedings{jiang2023vad,
  title     = {{VAD}: Vectorized Scene Representation for Efficient Autonomous Driving},
  author    = {Jiang, Bo and Chen, Shaoyu and Xu, Qing and Liao, Bencheng and
               Chen, Jiajie and Zhou, Helong and Zhang, Qian and Liu, Wenyu and
               Huang, Chang and Wang, Xinggang},
  booktitle = {Proceedings of the IEEE/CVF International Conference on Computer Vision},
  pages     = {8340--8350},
  year      = {2023}
}

@inproceedings{hu2023uniad,
  title     = {Planning-Oriented Autonomous Driving},
  author    = {Hu, Yihan and Yang, Jiazhi and Chen, Li and Li, Keyu and
               Sima, Chonghao and Zhu, Xizhou and Chai, Siqi and Du, Senyao and
               Lin, Tianwei and Wang, Wenhai and Lu, Lewei and Jia, Xiaosong and
               Liu, Qiang and Dai, Jifeng and Qiao, Yu and Li, Hongyang},
  booktitle = {Proceedings of the IEEE/CVF Conference on Computer Vision and Pattern Recognition},
  pages     = {17853--17862},
  year      = {2023}
}

@inproceedings{wang2020v2vnet,
  title     = {{V2VNet}: Vehicle-to-Vehicle Communication for Joint Perception and Prediction},
  author    = {Wang, Tsun-Hsuan and Manivasagam, Sivabalan and Liang, Ming and
               Yang, Bin and Zeng, Wenyuan and Urtasun, Raquel},
  booktitle = {Computer Vision -- ECCV 2020},
  pages     = {605--621},
  year      = {2020},
  publisher = {Springer},
  doi       = {10.1007/978-3-030-58536-5_36}
}

\end{document}